\documentclass[10pt,journal,compsoc]{IEEEtran}

\usepackage[numbers,sort&compress]{natbib}
\usepackage[pdftex]{graphicx}
\usepackage[cmex10]{amsmath}
\usepackage{hyperref}
\usepackage{url}
\usepackage{paralist}
\usepackage{booktabs}
\usepackage{color}
\usepackage{xcolor}
\usepackage{footmisc}

\graphicspath{{images/}}
\DeclareGraphicsExtensions{.png,.jpg,.pdf}

\newcommand{\ie}{\emph{i.e.},}
\newcommand{\eg}{\emph{e.g.},}
\newcommand{\vs}{\emph{vs.}}

\begin{document}

\title{Training on the test set? An analysis of \citet{spampinato2017}}

\author{Ren~Li,
  Jared~S.~Johansen,
  Hamad Ahmed,
  Thomas~V.~Ilyevsky,
  Ronnie~B~Wilbur,
  Hari~M~Bharadwaj,
  and~Jeffrey~Mark~Siskind,~\IEEEmembership{Senior~Member,~IEEE}%
  \thanks{All authors are with Purdue University, West Lafayette, IN 47907
    USA.
    R.~Li, J.~S.~Johansen, H.~Ahmed, T.~V.~Ilyevsky and J.~M.~Siskind are with
    the School of Electrical and Computer Engineering.
    R.~B.~Wilbur is with the the Department of Speech, Language, and Hearing
    Sciences and the Linguistics Program.
    H.~M.~Bharadwaj is with the Weldon School of Biomedical Engineering and the
    Department of Speech, Language, and Hearing Sciences.\protect\\
    E-mail: li2515@purdue.edu,
    jjohanse@purdue.edu,
    ahmed90@purdue.edu,
    tilyevsk@purdue.edu,
    wilbur@purdue.edu,
    hbharadwaj@purdue.edu,
    qobi@purdue.edu}%
  \thanks{Manuscript received}}

\markboth{IEEE TRANSACTIONS ON PATTERN ANALYSIS AND MACHINE INTELLIGENCE}%
{Li \etal\: Training on the test set? An analysis of \citet{spampinato2017}}

\IEEEtitleabstractindextext{%
\begin{abstract}
  A recent paper \citep{spampinato2017} claims to classify brain processing
  evoked in subjects watching ImageNet stimuli as measured with EEG and to
  employ a representation derived from this processing to construct a novel
  object classifier.
  That paper, together with a series of subsequent papers
  \citep{spampinato2016, palazzo2017, kavasidis2017, du2018, kumar2018,
    tirupattur2018, palazzo2018}, claims to revolutionize the field by
  achieving extremely successful results on a wide variety of computer-vision
  tasks, including object classification, transfer learning, and generation of
  images depicting human perception and thought using brain-derived
  representations measured through EEG.
  Our novel experiments and analyses demonstrate that their results crucially
  depend on the block design that they employ, where all stimuli of a given
  class are presented together, and fail with a rapid-event design, where
  stimuli of different classes are randomly intermixed.
  The block design leads to classification of arbitrary brain states based on
  block-level temporal correlations that tend to exist in all EEG data, rather
  than stimulus-related activity.
  Because every trial in their test sets comes from the same block as many
  trials in the corresponding training sets, their block design thus leads to
  surreptitiously training on the test set.
  This invalidates all subsequent analyses performed on this data in multiple
  published papers and calls into question all of the purported results.
  We further show that a novel object classifier constructed with a random
  codebook performs as well as or better than a novel object classifier
  constructed with the representation extracted from EEG data, suggesting that
  the performance of their classifier constructed with a representation
  extracted from EEG data does not benefit at all from the brain-derived
  representation.
  Our results calibrate the underlying difficulty of the tasks involved and
  caution against sensational and overly optimistic, but false, claims to the
  contrary.
\end{abstract}

}

\maketitle

\IEEEdisplaynontitleabstractindextext

\IEEEpeerreviewmaketitle

\IEEEraisesectionheading{\section{Introduction}\label{sec:introduction}}

\IEEEPARstart{A}{\hspace*{-5pt}} recent paper \citep{spampinato2017} claims to
(learn to) classify EEG data recorded from human subjects observing images from
ImageNet \citep{deng2009} and use the learned classifier to train a pure
computer-vision model.
In that paper, images from ImageNet are presented as stimuli to human subjects
undergoing EEG and a long short-term memory (LSTM \citep{hochreiter1997}),
combined with a fully connected layer and a ReLU layer, is trained to predict
the class of the stimulus from the recorded EEG signal.
The output of the ReLU layer is taken to reflect human neural encoding of the
percept.
The output of existing object classifiers is then regressed to this purported
human neural encoding of the percept in an attempt to have computer-vision
systems produce the same encoding of the percept.

That paper makes three specific claims \citep[\S~1 p.~6810]{spampinato2017}:
\begin{quote}
  \begin{compactenum}[1.]
  \item \emph{We propose a deep learning approach to classify EEG
    data evoked by visual object stimuli outperforming
    state-of-the-art methods both in the number of tackled
    object classes and in classification accuracy.}
    \label{enum:a}
  \item \emph{We propose the first computer vision approach driven
    by brain signals, \ie\ the first automated classification approach
    employing visual descriptors extracted directly from human neural
    processes involved in visual scene analysis.}
    \label{enum:b}
  \item \emph{We will publicly release the largest EEG dataset for visual object
    analysis, with related source code and trained models.}
    \label{enum:c}
  \end{compactenum}
\end{quote}
In particular, regarding claim~\ref{enum:a}, that paper further claims:
\begin{compactenum}[i.]
\item Their method can classify a far larger number (40) of distinct object
  classes than prior work (at most~12 \citep{kaneshiro2015}, typically~2) on
  classifying EEG signals.
\item Their method achieves far higher accuracy (82.9\%) than prior work
  \citep{kaneshiro2015} (29\%) on classifying EEG signals.
\end{compactenum}
That paper further couches its purported results in superlatives:
\begin{quote}
  \emph{In this paper, we want to take a great leap forward with
    respect to classic BCI approaches, \ie\ we aim at exploring a new and
    direct form of human involvement (a new vision of the ``human-based
    computation'' strategy) for automated visual classification.
    The underlying idea is to learn a brain signal discriminative manifold of
    visual categories by classifying EEG signals---reading the mind--and then to
    project images into such manifold to allow machines to perform automatic
    visual categorization---transfer human visual capabilities to machines.
    The impact of decoding object category-related EEG signals for inclusion
    into computer vision methods is tremendous.
    First, identifying EEG-based discriminative features for visual
    categorization might provide meaningful insight about the human visual
    perception systems.
    As a consequence, it will greatly advance performance of BCI-based
    applications as well as enable a new form of brain-based image
    labeling.
    Second, effectively projecting images into a new biologically based
    manifold will change radically the way object classifiers are developed
    (mainly in terms of feature extraction).}
  \citep[\S~1 pp.~6809--6810]{spampinato2017}
\end{quote}

Here, we report a number of experiments and analyses that call these results
and claims into question.
Specifically, we find that the classifier employed makes extensive, if not
sole, use of long-term static brain activity that persists much longer than the
duration of individual stimuli.
Since the paper employs a block design, where all stimuli of a given class are
presented to a subject in succession, the classifiers employed tend to classify
the brain activity during that block, which appears to be largely uncorrelated
with stimulus class.
This is exacerbated by the reliance of the classifier on DC and very-low
frequency (VLF) components in the EEG signal that reflect arbitrary long-term
static mental states during a block rather than dynamic brain activity.
Since each trial in the test sets employed comes from the same block as
many trials in the corresponding training sets, the reported high
classification accuracy results from surreptitious training on the test set.
When the experiment is repeated with a rapid-event design, where stimuli of
different classes are randomly intermixed, classification accuracy drops to
chance.
As a result, this renders suspect all of the results and claims advanced in
multiple published papers \citep{spampinato2016, spampinato2017, palazzo2017,
  kavasidis2017, du2018, kumar2018, tirupattur2018, palazzo2018}.
Our experiments suggest that the underlying tasks are far more difficult than
they appear on the surface and are far beyond the current state of the art.
This suggests caution in light of widely published \citep{spampinato2016,
  spampinato2017, palazzo2017, kavasidis2017, du2018, kumar2018,
  tirupattur2018, palazzo2018} sensational claims that are overly optimistic
but false.

\section{Overview}

In \S~\ref{sec:experiments}, we report a comprehensive set of experiments and
analyses to fully understand the results and claims reported by
\citet{spampinato2017}.
We first summarize our findings:
\begin{compactenum}[a.]
\item In \S~\ref{sec:reanalysis}, we reanalyze the EEG data collected by
  \citet{spampinato2017} using a number of different classifiers in addition to
  the one based on an LSTM that was employed by \citet{spampinato2017}.
  We show that one can obtain good, if not better, results with other
  classifiers, particularly ones that are sensitive to temporal alignment,
  unlike LSTMs.
  When we further reanalyze the EEG data collected by \citet{spampinato2017}
  with shorter temporal windows (as short as a single temporal sample), with
  random temporal offset, and with a reduced set of channels, we obtain even
  better results with these different classifiers.
  This suggests that the data collected by \citet{spampinato2017} lacks
  temporal and detailed spatial information reflective of perceptual processes
  that would benefit classification.
\item In \S~\ref{sec:newData}, we replicate the data collection effort of
  \citet{spampinato2017} using the same stimuli, presentation order, and
  timing, recording 96 channels with finer quantization (24 \vs\ 16 bits) and
  higher temporal sampling rate (4096~Hz \vs\ 1~kHz).
  We do this both with the original block design employed by
  \citet{spampinato2017}, where all stimuli of a given class are presented
  together, and with a rapid-event design, where stimuli of different classes
  are randomly intermixed.
  We also collect data with both the block and rapid-event designs,
  both for the original still-image stimuli depicting objects from ImageNet and
  short video clips depicting activity classes from Hollywood~2
  \citep{marszalek2009}.
\item In \S~\ref{sec:newAnalysis}, we replicate all of the analyses of
  \S~\ref{sec:reanalysis} on our new data.
  For data collected with the block design, we obtain moderately good
  classification accuracy on both image and video stimuli with one classifier,
  long windows, and a large set of channels.
  However, we obtain poor classification accuracy with all of the other
  classifiers, shorter windows, and a small set of channels.
  We further find that all classifiers yield chance performance on data
  collected with a rapid-event design.
\item \citet{spampinato2017} state that their data analysis included notch
  and bandpass filtering.
  Thus the analyses in \S~\ref{sec:newAnalysis} employed such filtering, which
  removes the DC and VLF components.
  Since \citet{palazzo2018a} and \citet{spampinato2018a} indicated to us in
  email (\S~\ref{sec:filtering}) that they did not perform bandpass
  filtering, in \S~\ref{sec:spectra}, we repeat the analysis of our data
  without bandpass filtering as well.
  Retaining the DC and VLF component allows us to replicate the results
  obtained on the data released by \citet{spampinato2017} with our data
  collected with a block design.
  However, we still obtain chance for our data collected with a rapid-event
  design.
\item The block design employed by \citet{spampinato2017}, together with
  their splits, has the property that every trial in each test set comes from
  a block that contains many trials in the corresponding training set.
  In \S~\ref{sec:design}, we conduct three new analyses.
  In the first new analysis, we repeat the analysis on the data released by
  \citet{spampinato2017} using new splits where the trials in each test set
  come from blocks that do not contain trials in the corresponding training
  set.
  Classification accuracy drops to chance.
  In the second new analysis, we repeat the analysis on our new data collected
  with a rapid-event design, where the labels are replaced with block-level
  labels instead of stimulus-level labels.
  Classification accuracy rises from chance to levels far above chance,
  reaching those obtained on the data collected by \citet{spampinato2017}.
  In the third new analysis, we rerun the code released by
  \citet{spampinato2017} on the data released by \citet{spampinato2017} after
  first applying various highpass filters to the data.
  Classification accuracy drops from roughly 93\% to roughly 32\%.
  Collectively these demonstrate that the high classification accuracies
  reported by \citet{spampinato2017} result from classifying the long-term
  brain activity associated with a block, even when that block contains stimuli
  of different classes, not the brain activity associated with perception of
  the class of the stimuli.
  They further demonstrate that this is exacerbated by the presence of DC and
  VLF components of the signal that remain due to lack of bandpass filtering.
  This refutes claims~\ref{enum:a} and~\ref{enum:c}.
\item In \S~\ref{sec:regression} and \S~\ref{sec:transfer}, we replicate
  the regression and transfer-learning analyses performed by \citet[\S~3.3,
    \S~4.2, and \S~4.3]{spampinato2017} but with a twist.
  We replace the EEG encodings with a random codebook and achieve equivalent,
  if not better, results.
  This demonstrates that the regression and transfer-learning analyses
  performed by \citet{spampinato2017} are not benefiting from a brain-inspired
  or brain-derived representation in any way, refuting claim~\ref{enum:b}.
\end{compactenum}

\section{Experiments}
\label{sec:experiments}

Our findings in \S~\ref{sec:discussion} and \S~\ref{sec:conclusion} are
supported by the following experiments and analyses performed.

\subsection{The \citet{spampinato2017} data collection}

\citet{spampinato2017} adopted the following experimental protocol.
They selected 40 object classes from ImageNet
\citep[footnote~1]{spampinato2017} along with 50 images for each class.
These were presented as stimuli to 6~human subjects undergoing EEG.\@
A block design was employed.
Each subject saw 40 blocks, each containing 50 image stimuli.
Each image was presented exactly once.
All 50 stimuli in a block were images of the same class.
All subjects saw exactly the same 2,000 images.
We do not know whether different subjects saw the classes, or the images in
a class, in different orders.
The image presentation order for one subject was provided to us by the authors.
Each image was presented for 0.5~s.
Blocks were separated by 10~s of blanking.
Approximately
$\textrm{40}\times(\textrm{50}\times\textrm{0.5~s}+\textrm{10~s})=\textrm{1400~s}$
of EEG data were collected from 128 channels at 1~kHz with 16~bit resolution.

\subsection{The \citet{spampinato2017} data analysis}

\citet{spampinato2017} report that the EEG data was preprocessed by application
of a notch filter (49--51~Hz) and a second-order band-pass Butterworth filter
(low cutoff frequency 14~Hz, high cut-off frequency 71~Hz).
The pass band was selected to include \emph{the Beta (15--31~Hz) and Gamma
(32--70~Hz) bands, as they convey information about the cognitive processes
involved in the visual perception} \citep[\S~3.1 p.~6812]{spampinato2017}.
The data for all 6~subjects was pooled, segmented into trials of
approximately 0.5~s duration, and divided into six
training/validation/test-set splits.
Each portion of each split contained data from all 6~subjects and all classes
for all subjects.
The data was z-scored prior to training and classification.
An LSTM, combined with a fully connected layer and a ReLU layer, was applied to
a 440~ms window of each trial starting 40~ms from stimulus onset.
A variety of different architectural parameters were evaluated, the best of
which achieved 85.4\% validation accuracy and 82.9\% test accuracy.
% omit [14]
\citet{spampinato2017} claim that this is significantly higher classification
accuracy for a significantly larger number of classes than all prior reported
classification experiments on EEG data \citep{cecotti2011, bashivan2015,
  stober2015, wang2012, carlson2011, simanova2010, bigdely2008, stewart2014,
  kaneshiro2015}.

\subsection{Reanalysis of the \citet{spampinato2017} data}
\label{sec:reanalysis}

We asked whether the significant improvement in classification ability was due
to the classifier architecture employed by \citet{spampinato2017} or whether it
was due to some aspect of their experimental protocol and data collection
procedure.
\citet{spampinato2017} have publicly released their
code\footnote{\url{http://perceive.dieei.unict.it/files/cvpr_2017_eeg_encoder.py}\label{foot:code}}
and
data.\footnote{\url{http://perceive.dieei.unict.it/index-dataset.php?name=EEG_Data}\label{foot:data}}.
This allowed us to verify their published results and to reanalyze their data
with different classifiers to investigate this question.
The released code yields (slightly better than) the published accuracy on the
released data.

\citet{spampinato2017} have released their data in both Python and Matlab
formats.
Both formats are subsequent to segmentation.
All results reported here were produced with the Python format data which was
z-scored before processing.
See \S~\ref{sec:reconciling} for details.

\begin{figure}
  \centering
  \includegraphics[width=0.75\columnwidth]{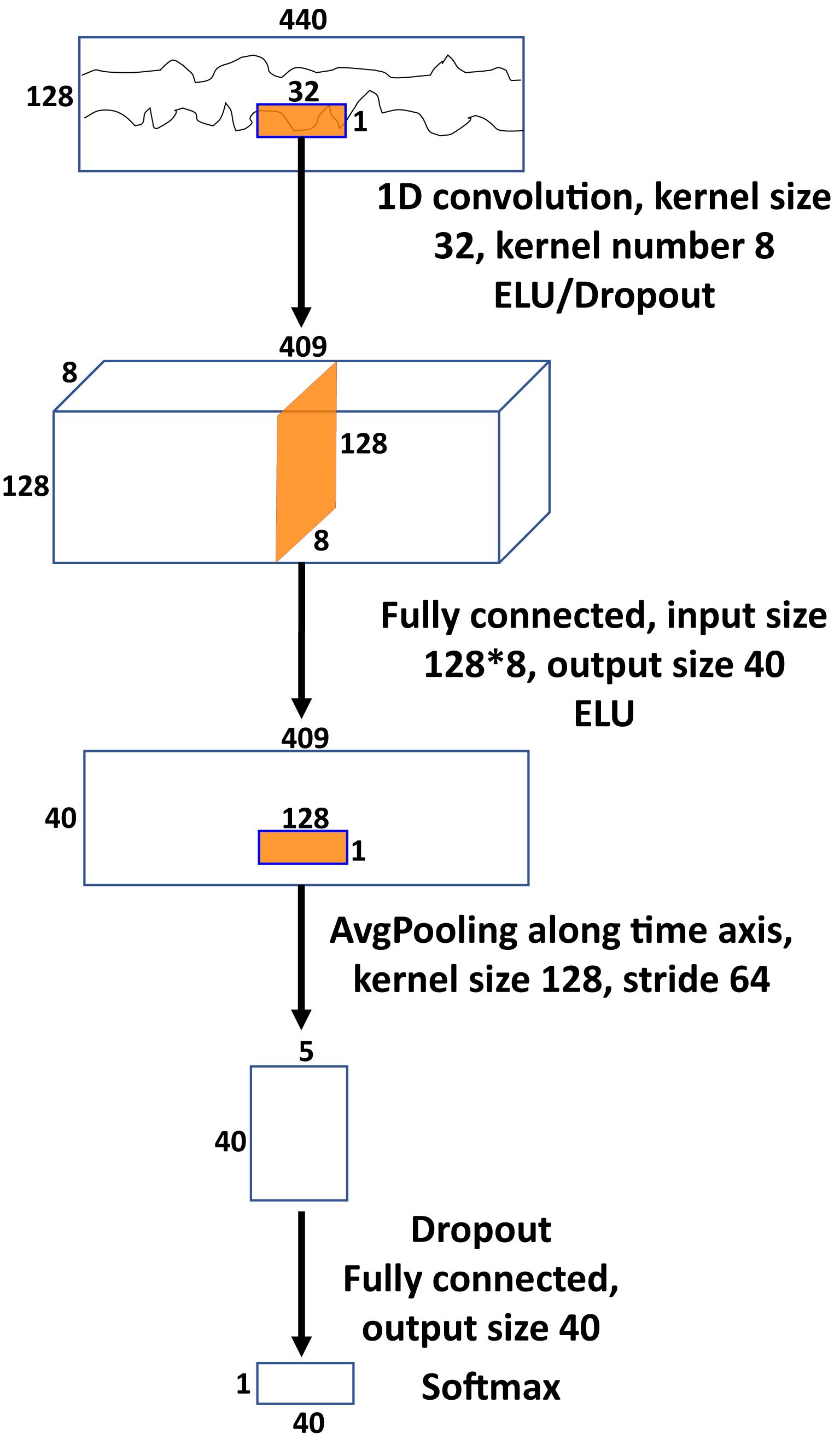}
  \caption{Our 1D~CNN used to process EEG data.}
  \label{fig:CNN}
\end{figure}

We reanalyzed the \citet{spampinato2017} data with four different classifiers
(Table~\ref{tab:classifiers}): a $k$-nearest neighbor classifier ($k$-NN), a
support vector machine (SVM \citep{cortes1995}), a multilayer perceptron (MLP),
and a 1D~convolutional neural network (CNN).\footnote{All code and data needed
  to replicate the results in this paper are available at
  \url{https://github.com/qobi/tpami2019}.}
The $k$-nearest-neighbor classifier used $k=7$ with a Euclidean distance on the
$\textrm{128}\times\textrm{440}=\textrm{56320}$ element vector associated with
each trial.
The SVM employed a linear kernel applied to data that was temporally
downsampled to 500 Hz, \ie\ $\textrm{128}\times\textrm{220}=\textrm{28160}$
element vectors.
The MLP employed two fully connected layers with a sigmoid activation function
after the first fully connected layer, and no dropout, trained with a
cross-entropy loss, applied to $\textrm{128}\times\textrm{440}=\textrm{56320}$
element vectors, with 128 hidden units.
The 1D~CNN (Fig.~\ref{fig:CNN}) processed each of the 128~channels
independently with eight 1D~CNNs of length~32 and stride~1.
The 128~applications of each of the eight 1D~CNNs shared the same parameters.
The output of each was processed by an ELU, followed by dropout with probability
of~0.5.
This yielded a temporal feature stream of length
$\textrm{440}-\textrm{32}+\textrm{1}=\textrm{409}$ with
$\textrm{128}\times\textrm{8}=\textrm{1024}$ features per time point.
This was then processed by a fully connected layer mapping each time point to a
40~element vector.
The parameters were shared across all time points.
This was then processed by average pooling along the time axis, independently
for each of the 40~channels, with a kernel of length~128 and a stride of~64.
%
% 0-127
% 64-191
% 128-255
% 192-319
% 256-282
%
This produced a feature map with 40~features for 5~time points.
Dropout with probability~0.5 was then applied, followed by a fully connected
layer with 40~outputs.
Training was performed with a cross-entropy loss.

\begin{table}
  \caption{Classification accuracy averaged across validation sets, test
    sets, and all 6~splits used by \citet{spampinato2017} on their released
    data with their software (an LSTM combined with a fully connected layer and
    a ReLU layer) and four  new classifiers: a nearest neighbor classifier
    ($k$-NN), an SVM, an MLP, and a 1D~CNN.}
  \label{tab:classifiers}
  \centering
  \resizebox{\columnwidth}{!}{\begin{tabular}{rrrrr} LSTM & $k$-NN & SVM &
        MLP & 1D~CNN\\ \hline 93.7\% & 42.9\% & 94.0\% & 49.0\% & 97.4\%
\end{tabular}} \end{table}

The results in Table~\ref{tab:classifiers} suggest that there is nothing
specific about the classifier architecture employed by \citet{spampinato2017}
that yields high results.
The same results can be obtained not only with an LSTM-based classifier or a
1D~CNN that attempts to model the temporal nature of the signal, but also with
an SVM that has no particular temporal structure.
Moreover, while other methods such as $k$-NN and MLP that also lack temporal
structure do not yield as high accuracy, they nonetheless yield accuracy far
higher than chance and far higher than any of the results reported in the
literature cited by \citet{spampinato2017}: \citep{cecotti2011, bashivan2015,
  stober2015, wang2012, carlson2011, simanova2010, bigdely2008, stewart2014,
  kaneshiro2015}.

Given that high accuracy was achieved with classifiers that should be sensitive
to temporal translation of the signal, we asked whether the classification
accuracy depended on this.
To this end, we trained and tested all 5~classifiers, varying the length of
the trial window between 200~ms, 100~ms, 50~ms, and 1~ms
(Table~\ref{tab:window}).\footnote{Due to the nature of its design, the 1D~CNN
  model was never applied to windows shorter than 200~ms.}
In all cases, the trial window was started at a random offset from the
stimulus onset, on a trial-by-trial basis.
Note that high accuracy can even be obtained with a single temporal sample
randomly selected within the stimulus interval.
This suggests that no temporal brain processing is reflected in the
classification accuracy.

\begin{table}
  \caption{Classification accuracy for varying trial window lengths with
    random temporal offset from the stimulus onset, averaged across validation
    sets, test sets, and all 6~splits used by \citet{spampinato2017} on their
    data with all 5~classifiers.}
  \label{tab:window}
  \centering
  \resizebox{\columnwidth}{!}{\begin{tabular}{r|rrrrr}
    window &   LSTM & $k$-NN &    SVM &    MLP & 1D~CNN\\
    \hline
    200~ms & 93.4\% & 39.5\% & 93.7\% & 62.9\% & 97.4\%\\
    100~ms & 95.1\% & 39.5\% & 93.5\% & 77.4\% &    n/a\\
    50~ms  & 96.3\% & 39.1\% & 93.7\% & 85.9\% &    n/a\\
    1~ms   & 92.9\% & 43.3\% & 90.3\% & 92.1\% &    n/a
  \end{tabular}}
\end{table}

An earlier report \citep{spampinato2016} conducted a similar data collection
effort to that of \citet{spampinato2017} with 32~channels instead of 128.
That effort yielded considerably lower classification accuracy (about 40\%) on
the same classes, stimuli, experimental protocol, and classification
architecture.
Given that the classifiers analyzed here appear not to rely on the temporal
nature of brain processing, we asked how much they rely on the number of
channels.
To this end, we performed feature \ie\ channel selection on the dataset to
train and test with various subsets of channels of different sizes.
The Fisher score \citep{gu2012} of a channel~$v$ for a classification task with
$C$~classes where each class~$c$ has~$n_c$ examples was computed as
\begin{equation}
  \frac{\displaystyle\sum_{c=1}^C n_c(\mu_{c,v}-\mu_v)^2}
       {\displaystyle\sum_{c=1}^C n_c \sigma_{c,v}^2}
\end{equation}
where~$\mu_{c,v}$ and~$\sigma_{c,v}$ were the per-class per-channel means and
variances and~$\mu_v$ was the per-channel mean.
We selected the~$m$ channels with highest Fisher score on the training set, for
varying~$m$, and repeated the training and testing on this subset of channels
for varying window lengths (Table~\ref{tab:channels}).
We observe that the full 128~channels are not necessary to achieve high
accuracy.
While the accuracy degrades somewhat when using fewer than 32~channels, one can
obtain far greater accuracy than chance and far greater accuracy than all prior
reported classification experiments cited by \citet{spampinato2017}:
\citep{cecotti2011, bashivan2015, stober2015, wang2012, carlson2011,
  simanova2010, bigdely2008, stewart2014, kaneshiro2015} on EEG data with as
few as 8~channels.
Moreover, one can obtain far greater accuracy than \citet{spampinato2016} with
the same number (32) of channels and the same accuracy with far fewer (8)
channels.
While the spatial layout of channel selection might not coincide with the
electrode placement of a cap with fewer electrodes, we next discuss why we
consider it important that one can accurately classify the
\citet{spampinato2017} data with such extreme spatial and temporal
downsampling.

\begin{table}
  \caption{Classification accuracy for varying numbers of channels, averaged
    across validation sets, test sets, and all 6~splits used by
    \citet{spampinato2017} on their data with all 5~classifiers and varying
    trial window lengths with random temporal offset from the stimulus onset.}
  \label{tab:channels}
  \centering
  \resizebox{\columnwidth}{!}{\begin{tabular}{r|r|rrrrr}
    window & channels &  LSTM  & $k$-NN &    SVM &    MLP & 1D~CNN\\
    \hline
    440~ms &       96 & 96.1\% & 45.4\% & 95.3\% & 62.7\% & 97.6\%\\
    200~ms &       96 & 95.4\% & 42.2\% & 94.7\% & 77.2\% & 97.8\%\\
    100~ms &       96 & 96.4\% & 42.7\% & 94.4\% & 84.8\% &    n/a\\
    50~ms  &       96 & 95.8\% & 41.3\% & 94.4\% & 90.6\% &    n/a\\
    1~ms   &       96 & 92.9\% & 49.5\% & 90.4\% & 93.0\% &    n/a\\
    \hline
    440~ms &       64 & 96.5\% & 55.4\% & 95.9\% & 76.8\% & 97.1\%\\
    200~ms &       64 & 96.4\% & 52.6\% & 95.0\% & 86.1\% & 97.5\%\\
    100~ms &       64 & 96.1\% & 52.4\% & 95.2\% & 90.3\% &    n/a\\
    50~ms  &       64 & 97.7\% & 52.5\% & 95.0\% & 93.3\% &    n/a\\
    1~ms   &       64 & 92.8\% & 61.8\% & 90.2\% & 93.2\% &    n/a\\
    \hline
    440~ms &       32 & 53.8\% & 58.4\% & 83.6\% & 85.6\% & 89.5\%\\
    200~ms &       32 & 91.1\% & 55.8\% & 81.0\% & 88.3\% & 90.2\%\\
    100~ms &       32 & 92.3\% & 55.8\% & 80.7\% & 90.3\% &    n/a\\
    50~ms  &       32 & 93.9\% & 55.1\% & 80.1\% & 90.5\% &    n/a\\
    1~ms   &       32 & 79.5\% & 62.3\% & 68.0\% & 81.6\% &    n/a\\
    \hline
    440~ms &       24 & 56.9\% & 56.4\% & 72.0\% & 82.7\% & 82.2\%\\
    200~ms &       24 & 80.0\% & 53.3\% & 67.7\% & 83.5\% & 82.3\%\\
    100~ms &       24 & 91.8\% & 53.4\% & 67.4\% & 84.7\% &    n/a\\
    50~ms  &       24 & 92.8\% & 52.7\% & 66.5\% & 86.6\% &    n/a\\
    1~ms   &       24 & 74.9\% & 58.7\% & 54.6\% & 73.0\% &    n/a\\
    \hline
    440~ms &       16 & 38.5\% & 57.1\% & 54.5\% & 78.7\% & 70.1\%\\
    200~ms &       16 & 57.5\% & 54.5\% & 49.6\% & 79.0\% & 68.9\%\\
    100~ms &       16 & 84.5\% & 54.1\% & 48.3\% & 79.4\% &    n/a\\
    50~ms  &       16 & 81.8\% & 53.3\% & 47.5\% & 79.2\% &    n/a\\
    1~ms   &       16 & 62.4\% & 55.8\% & 39.5\% & 61.5\% &    n/a\\
    \hline
    440~ms &        8 & 24.6\% & 48.1\% & 22.8\% & 57.8\% & 41.6\%\\
    200~ms &        8 & 45.0\% & 45.4\% & 18.6\% & 57.6\% & 41.4\%\\
    100~ms &        8 & 64.8\% & 44.5\% & 17.5\% & 57.9\% &    n/a\\
    50~ms  &        8 & 66.3\% & 44.0\% & 18.0\% & 59.5\% &    n/a\\
    1~ms   &        8 & 43.5\% & 42.9\% & 16.9\% & 37.9\% &    n/a\\
  \end{tabular}}
\end{table}

\subsection{New data collection}
\label{sec:newData}

The above analyses suggest that the accuracy achieved by \citet{spampinato2017}
was not due to the analysis architecture but rather due to either the
experimental protocol or the data collection effort.
We asked whether the accuracy was due to the former or the latter.
To this end, we repeated the data collection effort four times, all with the
same single subject.
The first two used the same 40~object classes and 2,000~image stimuli as
\citet{spampinato2017}.
The second two used the 12~activity classes and a subset of the video clips
from Hollywood~2 as described in \citet{siskind2015}.
The subset of clips was selected to be counterbalanced, with 32 clips per
class, temporally cropped to a uniform 4~s duration centered around the
activity class depicted, and transcoded to a uniform spatial and temporal
resolution.
Data was collected twice for each set of stimuli.
One collection used a block design, where all stimuli of a given class were
shown together in a single block.
The other collection used a rapid-event design, where the stimuli were
presented in randomized order.
The block design for the image stimuli employed the same design as
\citet{spampinato2017}: 40 blocks, each consisting of 50 stimuli, each
presented for 0.5~s with 10~s of blanking after each block.
The presentation order of the classes and stimuli within each class were the
same as in the data collected by \citet{spampinato2017}.
The rapid-event design for the image stimuli also employed 40 blocks, each
consisting of 50~stimuli, each presented for 0.5~s with 10~s of blanking after
each block, just that each block contained a random selection of images from
different classes.
In the latter, different blocks could contain different numbers of images for
different classes, subject to the constraint that, over the entire experiment,
each of the 2,000~images was shown exactly once.
The block design for the video stimuli began with 8~s of fixation blanking,
followed by 12~blocks, during each of which 32~clips were presented in
succession, each lasting 4~s, with 10~s of fixation blanking after each block.
%y
Approximately
$\textrm{12}\times(\textrm{32}\times\textrm{4~s}+\textrm{10~s})=\textrm{1656~s}$
of EEG data were collected.
For the block design, all stimuli within the block were of the same class.
The rapid-event design for the video stimuli also employed 12 blocks, each
consisting of 32~stimuli, each presented for 4~s with 10~s of blanking after
each block, just that each block contained a random selection of clips from
different classes.
In the latter, different blocks could contain different numbers of clips for
different classes, subject to the constraint that, over the entire experiment,
each of the 384~clips was shown exactly once.
Unlike the data collection effort of \citet{spampinato2017}, which divided each
recording into four 350~s sessions, each of our four recordings was collected
in a single session.

EEG data was recorded from 96 channels at 4096~Hz with 24~bit resolution using
a BioSemi ActiveTwo recorder\footnote{The ActiveTwo recorder employs 64$\times$
  oversampling and a sigma-delta A/D converter, yielding less quantization
  noise than 24 bit uniform sampling.} and a BioSemi gel electrode cap.
Two additional channels were used to record the signal from the earlobes for
rereferencing.
A trigger was recorded in the EEG data to indicate stimulus onset.
We downsampled the data to 1.024~kHz, rereferenced the data to the earlobes,
and employed the same preprocessing as reported by \citet{spampinato2017}: a
band-pass filter (low cutoff frequency 14~Hz, high cut-off frequency 71~Hz), a
notch filter (49--51~Hz), and z-scoring.\footnote{\citet{spampinato2017}
  presumably applied a notch filter to remove 50~Hz line noise.
  Being in the US, we should nominally remove 60~Hz line noise instead of 50~Hz.
  However, after rereferencing, our data contains no line noise so notch
  filtering is unnecessary.
  We employ a 50~Hz notch filter just to replicate \citet{spampinato2017}.}

\subsection{Analysis of our new data}
\label{sec:newAnalysis}

We applied the analysis from Table~\ref{tab:channels} to our new data collected
with the block design for the image (Table~\ref{tab:newDataBlockBandpass} left)
and video (Table~\ref{tab:newDataBlockBandpass} right) stimuli.
This subsumes all analyses performed on the \citet{spampinato2017} data.
Note that we are not able to replicate the results of \citet{spampinato2017}.
While the 1D~CNN achieves moderately good performance on both image and video
stimuli, the other classifiers perform poorly.
Moreover, for shorter analysis windows, random offsets, and reduced numbers of
channels, the other classifiers perform largely at chance.
We analyze the source of this difference below.

\begin{table*}
  \caption{Application of the analysis from Table~\protect\ref{tab:channels}
    to our new data collected with a block design on (left) image and (right)
    video stimuli, where our new data has been preprocessed with bandpass
    filtering.}
  \label{tab:newDataBlockBandpass}
  \centering
  \resizebox{\linewidth}{!}{
    \begin{tabular}{@{}c@{\hspace*{20pt}}c@{}}
      \begin{tabular}{r|r|rrrrr}
        window & channels &  LSTM  & $k$-NN &    SVM &    MLP & 1D~CNN\\
        \hline
        440~ms &       96 & 22.4\% &  2.5\% &  4.5\% &  4.9\% & 60.5\%\\
        200~ms &       96 & 20.5\% &  3.0\% &  3.3\% &  3.2\% & 53.8\%\\
        100~ms &       96 & 20.8\% &  3.2\% &  3.9\% &  3.7\% &    n/a\\
        50~ms  &       96 & 12.9\% &  3.7\% &  3.2\% &  3.8\% &    n/a\\
        1~ms   &       96 &  5.2\% &  5.3\% &  3.7\% &  3.8\% &    n/a\\
        \hline
        440~ms &       64 & 20.0\% &  2.9\% &  4.8\% &  4.1\% & 53.0\%\\
        200~ms &       64 & 11.2\% &  2.6\% &  3.0\% &  2.9\% & 40.7\%\\
        100~ms &       64 &  9.6\% &  3.3\% &  3.4\% &  3.3\% &    n/a\\
        50~ms  &       64 &  8.9\% &  3.8\% &  2.9\% &  2.9\% &    n/a\\
        1~ms   &       64 &  6.1\% &  4.4\% &  3.1\% &  4.3\% &    n/a\\
        \hline
        440~ms &       32 &  9.9\% &  3.3\% &  4.0\% &  3.7\% & 34.4\%\\
        200~ms &       32 &  8.4\% &  2.4\% &  3.3\% &  3.0\% & 24.7\%\\
        100~ms &       32 &  7.7\% &  2.6\% &  3.0\% &  2.5\% &    n/a\\
        50~ms  &       32 &  6.2\% &  3.3\% &  4.1\% &  3.4\% &    n/a\\
        1~ms   &       32 &  4.7\% &  4.2\% &  3.3\% &  3.4\% &    n/a\\
        \hline
        440~ms &       24 &  8.3\% &  3.1\% &  4.1\% &  3.6\% & 27.7\%\\
        200~ms &       24 &  7.9\% &  2.1\% &  3.4\% &  3.3\% & 27.3\%\\
        100~ms &       24 &  6.3\% &  2.2\% &  2.8\% &  2.5\% &    n/a\\
        50~ms  &       24 &  6.3\% &  2.8\% &  3.5\% &  2.2\% &    n/a\\
        1~ms   &       24 &  5.0\% &  4.4\% &  3.6\% &  3.8\% &    n/a\\
        \hline
        440~ms &       16 &  6.7\% &  2.4\% &  3.8\% &  3.2\% & 22.5\%\\
        200~ms &       16 &  6.9\% &  2.6\% &  2.5\% &  2.9\% & 19.0\%\\
        100~ms &       16 &  4.9\% &  2.5\% &  2.7\% &  3.1\% &    n/a\\
        50~ms  &       16 &  5.7\% &  2.6\% &  2.9\% &  3.2\% &    n/a\\
        1~ms   &       16 &  5.7\% &  3.7\% &  2.7\% &  3.7\% &    n/a\\
        \hline
        440~ms &        8 &  5.8\% &  2.9\% &  4.2\% &  3.5\% & 15.2\%\\
        200~ms &        8 &  5.2\% &  2.7\% &  3.1\% &  2.6\% & 13.8\%\\
        100~ms &        8 &  5.4\% &  2.7\% &  3.5\% &  2.5\% &    n/a\\
        50~ms  &        8 &  4.8\% &  3.5\% &  3.3\% &  3.2\% &    n/a\\
        1~ms   &        8 &  4.9\% &  3.6\% &  2.7\% &  3.7\% &    n/a\\
      \end{tabular}&
      \begin{tabular}{r|r|rrrrr}
        window  & channels &  LSTM  & $k$-NN &    SVM &    MLP & 1D~CNN\\
        \hline
        4000~ms &       96 & 13.3\% &  8.3\% &  8.3\% & 12.0\% & 65.9\%\\
        2000~ms &       96 & 15.6\% &  7.0\% & 10.4\% &  7.3\% & 66.7\%\\
        1000~ms &       96 & 14.8\% &  9.6\% &  8.9\% & 10.9\% & 64.3\%\\
        500~ms  &       96 & 12.2\% &  6.3\% &  7.0\% &  6.8\% &    n/a\\
        1~ms    &       96 & 13.8\% &  9.9\% &  9.1\% &  9.6\% &    n/a\\
        \hline
        4000~ms &       64 & 10.9\% &  7.3\% &  9.4\% & 10.9\% & 62.5\%\\
        2000~ms &       64 & 10.7\% &  7.8\% &  9.9\% &  9.1\% & 65.1\%\\
        1000~ms &       64 & 13.3\% &  7.8\% & 10.7\% &  8.3\% & 64.1\%\\
        500~ms  &       64 & 12.0\% &  7.8\% &  8.9\% &  9.6\% &    n/a\\
        1~ms    &       64 &  8.6\% &  9.9\% &  8.1\% & 10.1\% &    n/a\\
        \hline
        4000~ms &       32 &  9.4\% &  9.4\% & 13.5\% & 12.0\% & 51.6\%\\
        2000~ms &       32 & 12.5\% &  7.3\% &  8.1\% &  9.9\% & 61.7\%\\
        1000~ms &       32 & 12.2\% &  9.9\% & 10.7\% &  9.6\% & 52.6\%\\
        500~ms  &       32 & 12.0\% &  4.9\% &  8.1\% &  8.9\% &    n/a\\
        1~ms    &       32 & 14.8\% &  9.4\% & 11.2\% & 11.5\% &    n/a\\
        \hline
        4000~ms &       24 &  8.3\% &  7.3\% & 11.5\% &  9.9\% & 40.1\%\\
        2000~ms &       24 &  8.6\% &  7.8\% &  7.3\% &  7.6\% & 52.6\%\\
        1000~ms &       24 &  9.9\% &  9.9\% &  8.3\% &  7.6\% & 44.5\%\\
        500~ms  &       24 & 13.5\% &  6.8\% &  9.6\% &  9.6\% &    n/a\\
        1~ms    &       24 &  9.4\% &  6.3\% & 10.2\% &  9.6\% &    n/a\\
        \hline
        4000~ms &       16 &  9.6\% & 10.4\% & 10.4\% &  9.6\% & 44.8\%\\
        2000~ms &       16 & 11.5\% &  7.3\% &  7.3\% &  6.5\% & 47.9\%\\
        1000~ms &       16 &  7.8\% &  9.9\% &  9.6\% &  8.9\% & 41.1\%\\
        500~ms  &       16 & 11.2\% &  7.3\% & 10.7\% &  7.3\% &    n/a\\
        1~ms    &       16 &  9.4\% & 12.0\% &  6.0\% &  9.9\% &    n/a\\
        \hline
        4000~ms &        8 &  9.1\% &  8.3\% & 10.4\% &  9.1\% & 26.6\%\\
        2000~ms &        8 & 12.2\% &  8.6\% &  7.0\% & 10.4\% & 35.2\%\\
        1000~ms &        8 &  8.9\% &  7.0\% &  9.1\% &  6.8\% & 26.8\%\\
        500~ms  &        8 & 11.2\% &  8.3\% &  9.9\% &  8.9\% &    n/a\\
        1~ms    &        8 & 10.9\% & 12.2\% & 10.7\% & 10.4\% &    n/a\\
      \end{tabular}
  \end{tabular}}
\end{table*}

We then applied all of the classifiers from Table~\ref{tab:classifiers} to
the data collected with a rapid-event design for the image
(Table~\ref{tab:newDataRapidEventBandpass} left) and video
(Table~\ref{tab:newDataRapidEventBandpass} right) stimuli.
Note that all classifiers yield chance performance.

\begin{table*}
  \caption{Application of the analysis from Table~\protect\ref{tab:classifiers}
    to our new data collected with a rapid-event design on (left) image
    and (right) video stimuli, where our new data has been preprocessed with
    bandpass filtering.}
  \label{tab:newDataRapidEventBandpass}
  \centering
  \resizebox{\linewidth}{!}{\begin{tabular}{@{}c@{\hspace*{60pt}}c@{}}
    \begin{tabular}{rrrrr}
       LSTM & $k$-NN &    SVM &    MLP & 1D~CNN\\
      \hline
      2.9\% &  2.7\% &  3.4\% &  2.7\% &  2.1\%
    \end{tabular}&
    \begin{tabular}{rrrrr}
       LSTM & $k$-NN &    SVM &    MLP & 1D~CNN\\
      \hline
      8.3\% & 13.5\% &  7.3\% &  6.3\% & 10.4\%
    \end{tabular}
  \end{tabular}}
\end{table*}

\subsection{Spectral analysis}
\label{sec:spectra}

We asked why it is possible to achieve high accuracy with short analysis
windows on the \citet{spampinato2017} data but not with our data.
\citet{palazzo2018a} and \citet{spampinato2018a} indicated to us in email that
their report of preprocessing was a misprint and that they performed notch
filtering (during acquisition) and z-scoring but not bandpass filtering.
See \S~\ref{sec:filtering} for details.
Since their released code performs z-scoring, this indicates that their
released data reflects notch filtering but neither bandpass filtering nor
z-scoring.
We thus reanalyzed our data with a notch filter and z-scoring but no bandpass
filter (Tables~\ref{tab:newDataBlockNoBandpass}
and~\ref{tab:newDataRapidEventNoBandpass}).
Note that we now obtain better results for the data collected with the block
design, similar to that obtained with the data released by
\citet{spampinato2017}, but still obtain chance for data collected with the
rapid-event design.

\begin{table*}
  \caption{Application of the analysis from Table~\protect\ref{tab:channels}
    to our new data collected with a block design on (left) image and (right)
    video stimuli, where our new data has not been preprocessed with bandpass
    filtering.}
  \label{tab:newDataBlockNoBandpass}
  \centering
  \resizebox{\linewidth}{!}{
    \begin{tabular}{@{}c@{\hspace*{20pt}}c@{}}
      \begin{tabular}{r|r|rrrrr}
        window & channels &  LSTM  & $k$-NN &    SVM &    MLP & 1D~CNN\\
        \hline
        440~ms &       96 & 63.1\% &100.0\% &100.0\% & 21.9\% & 85.9\%\\
        200~ms &       96 & 66.1\% & 99.9\% &100.0\% & 35.8\% & 83.7\%\\
        100~ms &       96 & 71.8\% &100.0\% &100.0\% & 45.7\% &    n/a\\
        50~ms  &       96 & 70.2\% & 99.8\% &100.0\% & 63.3\% &    n/a\\
        1~ms   &       96 & 59.7\% & 99.9\% & 99.8\% & 81.3\% &    n/a\\
        \hline
        440~ms &       64 & 70.6\% & 99.9\% &100.0\% & 27.5\% & 76.4\%\\
        200~ms &       64 & 64.3\% &100.0\% &100.0\% & 34.1\% & 78.2\%\\
        100~ms &       64 & 68.1\% & 99.9\% &100.0\% & 50.5\% &    n/a\\
        50~ms  &       64 & 70.9\% & 99.8\% &100.0\% & 68.6\% &    n/a\\
        1~ms   &       64 & 58.1\% &100.0\% & 99.5\% & 71.3\% &    n/a\\
        \hline
        440~ms &       32 & 64.1\% & 99.8\% &100.0\% & 31.5\% & 74.7\%\\
        200~ms &       32 & 60.7\% & 99.7\% &100.0\% & 43.9\% & 73.9\%\\
        100~ms &       32 & 63.1\% & 99.8\% &100.0\% & 62.1\% &    n/a\\
        50~ms  &       32 & 66.4\% & 99.7\% &100.0\% & 73.0\% &    n/a\\
        1~ms   &       32 & 45.6\% & 99.7\% & 99.0\% & 63.8\% &    n/a\\
        \hline
        440~ms &       24 & 48.4\% & 99.9\% &100.0\% & 35.5\% & 73.3\%\\
        200~ms &       24 & 54.6\% & 99.7\% &100.0\% & 51.5\% & 71.8\%\\
        100~ms &       24 & 59.4\% & 99.8\% &100.0\% & 69.8\% &    n/a\\
        50~ms  &       24 & 71.1\% & 99.9\% &100.0\% & 80.0\% &    n/a\\
        1~ms   &       24 & 50.7\% & 99.7\% & 98.7\% & 62.2\% &    n/a\\
        \hline
        440~ms &       16 & 51.5\% & 99.9\% &100.0\% & 40.1\% & 70.2\%\\
        200~ms &       16 & 56.9\% & 99.8\% &100.0\% & 55.8\% & 72.5\%\\
        100~ms &       16 & 63.2\% & 99.7\% &100.0\% & 71.3\% &    n/a\\
        50~ms  &       16 & 63.8\% & 99.8\% &100.0\% & 70.0\% &    n/a\\
        1~ms   &       16 & 45.3\% & 99.6\% & 98.8\% & 62.7\% &    n/a\\
        \hline
        440~ms &        8 & 31.8\% & 99.6\% & 99.9\% & 53.9\% & 71.5\%\\
        200~ms &        8 & 40.4\% & 99.5\% &100.0\% & 63.4\% & 71.3\%\\
        100~ms &        8 & 54.2\% & 99.5\% & 99.8\% & 66.2\% &    n/a\\
        50~ms  &        8 & 61.2\% & 99.4\% & 99.3\% & 61.9\% &    n/a\\
        1~ms   &        8 & 46.0\% & 99.5\% & 98.7\% & 58.8\% &    n/a\\
      \end{tabular}&
      \begin{tabular}{r|r|rrrrr}
        window  & channels &  LSTM  & $k$-NN &    SVM &    MLP & 1D~CNN\\
        \hline
        4000~ms &       96 & 98.4\% & 97.9\% & 99.0\% & 77.6\% & 99.0\%\\
        2000~ms &       96 & 94.8\% & 98.2\% & 99.5\% & 85.4\% & 99.0\%\\
        1000~ms &       96 & 96.9\% & 97.4\% & 99.0\% & 95.6\% & 99.0\%\\
        500~ms  &       96 & 88.5\% & 96.4\% & 99.5\% & 98.7\% &    n/a\\
        1~ms    &       96 & 91.7\% & 97.4\% & 99.2\% & 98.7\% &    n/a\\
        \hline
        4000~ms &       64 & 96.9\% & 96.9\% &100.0\% & 68.2\% & 98.7\%\\
        2000~ms &       64 & 95.8\% & 97.1\% &100.0\% & 83.1\% & 99.0\%\\
        1000~ms &       64 & 94.3\% & 97.1\% &100.0\% & 95.6\% & 98.4\%\\
        500~ms  &       64 & 89.8\% & 96.6\% &100.0\% & 99.0\% &    n/a\\
        1~ms    &       64 & 78.6\% & 97.9\% &100.0\% & 97.9\% &    n/a\\
        \hline
        4000~ms &       32 & 91.9\% & 96.9\% &100.0\% & 75.0\% & 99.7\%\\
        2000~ms &       32 & 77.1\% & 96.9\% &100.0\% & 86.2\% & 97.9\%\\
        1000~ms &       32 & 85.7\% & 96.6\% &100.0\% & 96.4\% & 98.7\%\\
        500~ms  &       32 & 82.6\% & 96.1\% & 99.7\% & 99.2\% &    n/a\\
        1~ms    &       32 & 76.8\% & 97.9\% &100.0\% & 96.9\% &    n/a\\
        \hline
        4000~ms &       24 & 84.1\% & 96.9\% &100.0\% & 83.3\% & 99.0\%\\
        2000~ms &       24 & 83.9\% & 97.1\% &100.0\% & 84.1\% & 99.0\%\\
        1000~ms &       24 & 69.3\% & 97.1\% &100.0\% & 95.3\% & 98.7\%\\
        500~ms  &       24 & 80.2\% & 96.9\% & 99.7\% & 97.9\% &    n/a\\
        1~ms    &       24 & 60.2\% & 95.6\% & 99.5\% & 95.8\% &    n/a\\
        \hline
        4000~ms &       16 & 75.5\% & 96.9\% &100.0\% & 69.8\% & 97.7\%\\
        2000~ms &       16 & 74.7\% & 97.4\% &100.0\% & 85.4\% & 95.6\%\\
        1000~ms &       16 & 63.5\% & 97.1\% &100.0\% & 94.0\% & 98.4\%\\
        500~ms  &       16 & 71.1\% & 97.7\% &100.0\% & 96.9\% &    n/a\\
        1~ms    &       16 & 63.8\% & 97.7\% & 99.7\% & 91.1\% &    n/a\\
        \hline
        4000~ms &        8 & 63.0\% & 97.9\% &100.0\% & 80.2\% & 98.2\%\\
        2000~ms &        8 & 62.0\% & 99.0\% & 99.7\% & 89.3\% & 99.2\%\\
        1000~ms &        8 & 78.9\% & 99.2\% &100.0\% & 93.5\% & 98.4\%\\
        500~ms  &        8 & 72.7\% & 99.0\% &100.0\% & 96.9\% &    n/a\\
        1~ms    &        8 & 55.7\% & 98.7\% & 99.5\% & 90.4\% &    n/a\\
      \end{tabular}
  \end{tabular}}
\end{table*}

\begin{table*}
  \caption{Application of the analysis from Table~\protect\ref{tab:classifiers}
    to our new data collected with a rapid-event design on (left) image
    and (right) video stimuli, where our new data has not been preprocessed with
    bandpass filtering.}
  \label{tab:newDataRapidEventNoBandpass}
  \centering
  \resizebox{\linewidth}{!}{\begin{tabular}{@{}c@{\hspace*{60pt}}c@{}}
    \begin{tabular}{rrrrr}
       LSTM & $k$-NN &    SVM &    MLP & 1D~CNN\\
      \hline
      0.7\% &  1.4\% &  2.7\% &  1.5\% &  2.1\%
    \end{tabular}&
    \begin{tabular}{rrrrr}
       LSTM & $k$-NN &    SVM &    MLP & 1D~CNN\\
      \hline
     10.2\% &  8.3\% &  7.3\% &  9.6\% & 10.7\%
    \end{tabular}
  \end{tabular}}
\end{table*}

\subsection{Block \vs\ Rapid-Event Design}
\label{sec:design}

We asked why we (and \citet{spampinato2017}) are able to obtain high
classification accuracy with a block design but not a rapid-event design.
To this end, we performed three reanalyses.
First, we repeated the analysis from
Tables~\ref{tab:classifiers}--\ref{tab:channels}, where instead of using the
training/test set splits provided by \citet{spampinato2017}, we conducted a
leave-one-subject-out round-robin cross validation, training on all data from
five of the subjects and testing on all data from the sixth, rotating among all
six subjects as test (Table~\ref{tab:reanalysisA}).
Note that classification accuracy is now at chance.

\begin{table}
  \caption{Reanalysis of the data released by \citet{spampinato2017} with
    classification accuracy averaged over leave-one-subject-out round-robin
    cross validation instead of the provided splits.}
  \label{tab:reanalysisA}
  \centering
  \resizebox{\columnwidth}{!}{\begin{tabular}{r|r|rrrrr}
    window & channels &  LSTM  & $k$-NN &    SVM &    MLP & 1D~CNN\\
    \hline
    440~ms &      128 &  2.7\% &  3.4\% &  3.4\% &  3.4\% &  3.1\%\\
    200~ms &      128 &  2.4\% &  3.5\% &  3.3\% &  3.5\% &  3.1\%\\
    100~ms &      128 &  3.0\% &  3.4\% &  3.5\% &  2.6\% &    n/a\\
    50~ms  &      128 &  2.9\% &  3.3\% &  3.0\% &  2.4\% &    n/a\\
    1~ms   &      128 &  3.0\% &  2.2\% &  3.1\% &  3.0\% &    n/a\\
    \hline
    440~ms &       96 &  3.3\% &  3.0\% &  3.4\% &  2.6\% &  3.3\%\\
    200~ms &       96 &  2.3\% &  3.0\% &  3.3\% &  3.6\% &  4.0\%\\
    100~ms &       96 &  2.8\% &  2.6\% &  3.1\% &  3.5\% &    n/a\\
    50~ms  &       96 &  2.9\% &  2.6\% &  3.2\% &  2.5\% &    n/a\\
    1~ms   &       96 &  2.7\% &  2.4\% &  3.2\% &  3.2\% &    n/a\\
    \hline
    440~ms &       64 &  2.6\% &  2.3\% &  3.4\% &  3.6\% &  3.6\%\\
    200~ms &       64 &  3.2\% &  2.2\% &  3.0\% &  2.7\% &  3.2\%\\
    100~ms &       64 &  2.7\% &  2.2\% &  3.2\% &  3.2\% &    n/a\\
    50~ms  &       64 &  2.6\% &  2.0\% &  2.9\% &  3.2\% &    n/a\\
    1~ms   &       64 &  2.7\% &  2.4\% &  3.0\% &  3.0\% &    n/a\\
    \hline
    440~ms &       32 &  2.6\% &  1.9\% &  3.7\% &  2.6\% &  3.6\%\\
    200~ms &       32 &  3.0\% &  1.8\% &  3.9\% &  3.4\% &  2.7\%\\
    100~ms &       32 &  3.4\% &  1.9\% &  3.8\% &  2.6\% &    n/a\\
    50~ms  &       32 &  2.6\% &  2.1\% &  4.2\% &  3.2\% &    n/a\\
    1~ms   &       32 &  1.8\% &  2.3\% &  2.8\% &  3.2\% &    n/a\\
    \hline
    440~ms &       24 &  2.3\% &  2.3\% &  3.7\% &  2.6\% &  2.9\%\\
    200~ms &       24 &  2.5\% &  2.5\% &  3.6\% &  2.7\% &  2.7\%\\
    100~ms &       24 &  3.1\% &  2.4\% &  3.6\% &  2.4\% &    n/a\\
    50~ms  &       24 &  3.0\% &  2.8\% &  3.6\% &  2.5\% &    n/a\\
    1~ms   &       24 &  2.9\% &  3.0\% &  3.0\% &  2.5\% &    n/a\\
    \hline
    440~ms &       16 &  2.0\% &  2.1\% &  2.5\% &  2.7\% &  2.8\%\\
    200~ms &       16 &  1.9\% &  2.1\% &  2.7\% &  2.7\% &  2.7\%\\
    100~ms &       16 &  2.7\% &  2.2\% &  2.3\% &  2.6\% &    n/a\\
    50~ms  &       16 &  2.6\% &  2.4\% &  2.9\% &  2.5\% &    n/a\\
    1~ms   &       16 &  3.1\% &  2.6\% &  3.2\% &  2.8\% &    n/a\\
    \hline
    440~ms &        8 &  2.1\% &  2.3\% &  2.8\% &  3.0\% &  2.9\%\\
    200~ms &        8 &  2.8\% &  2.3\% &  2.5\% &  2.8\% &  2.4\%\\
    100~ms &        8 &  2.6\% &  2.3\% &  2.5\% &  3.1\% &    n/a\\
    50~ms  &        8 &  1.9\% &  2.4\% &  2.8\% &  2.8\% &    n/a\\
    1~ms   &        8 &  2.8\% &  2.2\% &  3.0\% &  2.7\% &    n/a\\
  \end{tabular}}
\end{table}

Second, we reran all of the analyses from
Table~\ref{tab:newDataBlockNoBandpass} on our new data collected with a
rapid-event design, both with and without bandpass filtering, but with a twist.
Instead of using correct labels, which varied on a stimulus-by-stimulus basis,
we used random labels, which varied on a block-by-block basis: each block was
given a distinct label but all stimuli within a block were given the same label.
Thus while the stimuli are changing in each block, they are given the wrong
unchanging label and, like the block design employed by \citet{spampinato2017},
each trial in the test set comes from a block with many trials in the
training set.
The results with and without bandpass filtering are shown in
Tables~\ref{tab:reanalysisBbandpass} and~\ref{tab:reanalysisBnoBandpass}
respectively and mirror the results in Tables~\ref{tab:newDataBlockBandpass}
and~\ref{tab:newDataBlockNoBandpass} respectively.
Note that with bandpass filtering, we obtain classification accuracies far
higher than chance with the 1D~CNN, while without bandpass filtering, we obtain
near perfect classification accuracies, similar to those obtained in
Tables~~\ref{tab:classifiers}--\ref{tab:channels}.

\begin{table*}
  \caption{Reanalysis of our new data collected with a rapid-event design on
    (left) image and (right) video stimuli with incorrect block-level labels,
    where our new data has been preprocessed with bandpass filtering.}
  \label{tab:reanalysisBbandpass}
  \centering
  \resizebox{\linewidth}{!}{
    \begin{tabular}{@{}c@{\hspace*{20pt}}c@{}}
      \begin{tabular}{r|r|rrrrr}
        window & channels &  LSTM  & $k$-NN &    SVM &    MLP & 1D~CNN\\
        \hline
	440~ms &       96 &  9.2\% &  2.3\% &  2.7\% &  3.1\% & 47.7\%\\
	200~ms &       96 &  8.8\% &  2.3\% &  2.6\% &  3.2\% & 38.3\%\\
	100~ms &       96 &  6.9\% &  2.8\% &  3.5\% &  2.5\% &    n/a\\
	50~ms  &       96 &  6.3\% &  2.9\% &  3.3\% &  3.2\% &    n/a\\
	1~ms   &       96 &  3.1\% &  2.5\% &  2.5\% &  2.6\% &    n/a\\
	\hline
	440~ms &       64 &  6.3\% &  2.4\% &  2.0\% &  2.4\% & 34.4\%\\
	200~ms &       64 &  4.1\% &  2.6\% &  2.3\% &  2.9\% & 28.3\%\\
	100~ms &       64 &  4.0\% &  2.0\% &  2.5\% &  2.1\% &    n/a\\
	50~ms  &       64 &  3.9\% &  2.1\% &  2.5\% &  2.8\% &    n/a\\
	1~ms   &       64 &  3.1\% &  2.2\% &  3.0\% &  3.8\% &    n/a\\
	\hline
	440~ms &       32 &  3.6\% &  2.1\% &  2.0\% &  2.3\% & 17.7\%\\
	200~ms &       32 &  2.7\% &  2.7\% &  2.5\% &  2.4\% & 13.8\%\\
	100~ms &       32 &  2.1\% &  2.5\% &  2.6\% &  2.4\% &    n/a\\
	50~ms  &       32 &  2.4\% &  2.2\% &  2.5\% &  3.3\% &    n/a\\
	1~ms   &       32 &  3.5\% &  3.1\% &  2.4\% &  3.2\% &    n/a\\
	\hline
	440~ms &       24 &  2.9\% &  1.7\% &  2.9\% &  2.8\% & 15.0\%\\
	200~ms &       24 &  3.8\% &  2.4\% &  2.5\% &  1.8\% & 15.4\%\\
	100~ms &       24 &  4.0\% &  2.1\% &  2.8\% &  2.6\% &    n/a\\
	50~ms  &       24 &  2.6\% &  2.1\% &  2.9\% &  2.9\% &    n/a\\
	1~ms   &       24 &  3.6\% &  2.7\% &  2.5\% &  3.0\% &    n/a\\
	\hline
	440~ms &       16 &  2.5\% &  1.7\% &  2.9\% &  2.7\% & 12.8\%\\
	200~ms &       16 &  3.3\% &  2.4\% &  3.0\% &  2.7\% & 10.0\%\\
	100~ms &       16 &  2.7\% &  2.3\% &  2.2\% &  2.0\% &    n/a\\
	50~ms  &       16 &  2.5\% &  2.9\% &  3.3\% &  2.4\% &    n/a\\
	1~ms   &       16 &  3.6\% &  3.0\% &  2.3\% &  2.2\% &    n/a\\
	\hline
	440~ms &        8 &  2.1\% &  1.6\% &  2.5\% &  2.7\% &  6.9\%\\
	200~ms &        8 &  3.0\% &  2.0\% &  2.7\% &  2.4\% &  9.7\%\\
	100~ms &        8 &  3.4\% &  1.8\% &  3.1\% &  2.3\% &    n/a\\
	50~ms  &        8 &  2.4\% &  2.5\% &  2.4\% &  2.4\% &    n/a\\
	1~ms   &        8 &  2.7\% &  3.1\% &  2.1\% &  2.4\% &    n/a\\
      \end{tabular}&
      \begin{tabular}{r|r|rrrrr}
        window  & channels &  LSTM  & $k$-NN &    SVM &    MLP & 1D~CNN\\
        \hline
        4000~ms &       96 & 19.8\% &  7.3\% &  8.3\% &  6.5\% & 72.9\%\\
        2000~ms &       96 & 18.8\% &  8.1\% &  8.9\% &  7.3\% & 74.2\%\\
        1000~ms &       96 & 20.3\% &  8.1\% &  7.8\% &  8.1\% & 70.1\%\\
        500~ms  &       96 & 15.9\% &  9.4\% &  9.4\% &  6.5\% &    n/a\\
        1~ms    &       96 &  9.6\% &  9.9\% &  8.1\% &  9.4\% &    n/a\\
        \hline
        4000~ms &       64 & 16.9\% &  8.3\% &  6.3\% &  6.8\% & 66.7\%\\
        2000~ms &       64 & 13.0\% &  8.1\% &  9.1\% &  8.3\% & 66.1\%\\
        1000~ms &       64 & 15.9\% &  7.3\% &  7.3\% &  7.8\% & 60.2\%\\
        500~ms  &       64 & 13.5\% &  8.6\% &  8.3\% &  8.3\% &    n/a\\
        1~ms    &       64 & 10.9\% &  9.1\% &  9.6\% &  8.9\% &    n/a\\
        \hline
        4000~ms &       32 & 11.5\% &  6.3\% &  6.3\% &  7.6\% & 42.7\%\\
        2000~ms &       32 &  9.9\% &  9.4\% &  8.9\% &  7.3\% & 44.5\%\\
        1000~ms &       32 &  8.6\% &  9.9\% & 10.2\% & 12.2\% & 35.4\%\\
        500~ms  &       32 & 14.1\% &  9.1\% &  7.3\% &  8.1\% &    n/a\\
        1~ms    &       32 & 10.4\% & 10.4\% & 10.2\% &  9.6\% &    n/a\\
        \hline
        4000~ms &       24 & 18.0\% &  4.2\% &  6.3\% &  7.0\% & 46.1\%\\
        2000~ms &       24 & 12.0\% &  7.8\% &  9.1\% &  7.3\% & 32.0\%\\
        1000~ms &       24 & 12.8\% &  8.9\% & 10.2\% &  7.6\% & 44.3\%\\
        500~ms  &       24 &  7.6\% &  7.8\% & 10.2\% &  5.7\% &    n/a\\
        1~ms    &       24 & 10.2\% &  9.1\% &  5.7\% &  7.8\% &    n/a\\
        \hline
        4000~ms &       16 & 15.6\% &  7.3\% &  4.2\% &  7.0\% & 49.5\%\\
        2000~ms &       16 & 14.6\% &  9.1\% &  7.3\% &  9.9\% & 37.8\%\\
        1000~ms &       16 & 10.7\% &  8.6\% & 10.2\% & 11.2\% & 36.2\%\\
        500~ms  &       16 & 13.0\% &  6.5\% &  8.3\% &  8.1\% &    n/a\\
        1~ms    &       16 &  7.8\% &  9.6\% &  7.3\% &  7.8\% &    n/a\\
        \hline
        4000~ms &        8 & 10.2\% &  9.4\% &  8.3\% &  8.6\% & 28.1\%\\
        2000~ms &        8 & 13.3\% &  8.9\% &  8.1\% &  7.6\% & 40.3\%\\
        1000~ms &        8 &  9.6\% &  8.3\% &  9.9\% & 11.5\% & 21.4\%\\
        500~ms  &        8 &  9.9\% &  8.1\% &  7.8\% &  7.3\% &    n/a\\
        1~ms    &        8 &  7.6\% &  8.9\% &  9.1\% & 10.2\% &    n/a\\
      \end{tabular}
  \end{tabular}}
\end{table*}

\begin{table*}
  \caption{Reanalysis of our new data collected with a rapid-event design on
    (left) image and (right) video stimuli with incorrect block-level labels,
    where our new data has not been preprocessed with bandpass filtering.}
  \label{tab:reanalysisBnoBandpass}
  \centering
  \resizebox{\linewidth}{!}{
    \begin{tabular}{@{}c@{\hspace*{20pt}}c@{}}
      \begin{tabular}{r|r|rrrrr}
        window & channels &  LSTM  & $k$-NN &    SVM &    MLP & 1D~CNN\\
        \hline
        440~ms &       96 & 76.7\% & 99.8\% &100.0\% & 37.8\% & 95.4\%\\
        200~ms &       96 & 74.3\% & 99.8\% &100.0\% & 53.5\% & 98.6\%\\
        100~ms &       96 & 78.8\% & 99.8\% &100.0\% & 65.0\% &    n/a\\
        50~ms  &       96 & 80.7\% & 99.9\% &100.0\% & 83.9\% &    n/a\\
        1~ms   &       96 & 58.4\% & 99.9\% &100.0\% & 98.8\% &    n/a\\
        \hline
        440~ms &       64 & 66.3\% & 99.8\% &100.0\% & 29.9\% & 96.3\%\\
        200~ms &       64 & 67.6\% & 99.8\% &100.0\% & 36.1\% & 96.3\%\\
        100~ms &       64 & 72.9\% & 99.9\% &100.0\% & 61.7\% &    n/a\\
        50~ms  &       64 & 75.0\% & 99.8\% &100.0\% & 89.5\% &    n/a\\
        1~ms   &       64 & 962.1\% & 99.9\% &100.0\% & 93.0\% &    n/a\\
        \hline
        440~ms &       32 & 51.4\% & 99.3\% &100.0\% & 31.3\% & 84.9\%\\
        200~ms &       32 & 62.2\% & 99.5\% &100.0\% & 46.7\% & 82.4\%\\
        100~ms &       32 & 65.6\% & 99.4\% &100.0\% & 73.5\% &    n/a\\
        50~ms  &       32 & 64.8\% & 99.5\% &100.0\% & 86.6\% &    n/a\\
        1~ms   &       32 & 49.0\% & 99.2\% & 97.9\% & 71.2\% &    n/a\\
        \hline
        440~ms &       24 & 50.7\% & 99.4\% &100.0\% & 36.1\% & 81.4\%\\
        200~ms &       24 & 60.0\% & 99.4\% &100.0\% & 60.0\% & 80.8\%\\
        100~ms &       24 & 55.3\% & 99.4\% &100.0\% & 76.7\% &    n/a\\
        50~ms  &       24 & 63.3\% & 99.3\% & 99.9\% & 82.9\% &    n/a\\
        1~ms   &       24 & 50.9\% & 99.3\% & 97.8\% & 70.5\% &    n/a\\
        \hline
        440~ms &       16 & 46.9\% & 99.3\% &100.0\% & 45.0\% & 80.5\%\\
        200~ms &       16 & 53.6\% & 99.2\% &100.0\% & 66.0\% & 76.5\%\\
        100~ms &       16 & 63.9\% & 99.0\% &100.0\% & 80.9\% &    n/a\\
        50~ms  &       16 & 63.1\% & 99.0\% & 99.8\% & 84.6\% &    n/a\\
        1~ms   &       16 & 43.0\% & 98.8\% & 96.8\% & 64.1\% &    n/a\\
        \hline
        440~ms &        8 & 37.8\% & 98.4\% & 99.4\% & 60.3\% & 69.0\%\\
        200~ms &        8 & 41.4\% & 98.5\% & 99.6\% & 71.2\% & 70.9\%\\
        100~ms &        8 & 52.6\% & 98.5\% & 99.2\% & 75.5\% &    n/a\\
        50~ms  &        8 & 54.3\% & 98.1\% & 98.8\% & 75.8\% &    n/a\\
        1~ms   &        8 & 44.2\% & 98.3\% & 95.0\% & 55.5\% &    n/a\\
      \end{tabular}&
      \begin{tabular}{r|r|rrrrr}
        window  & channels &  LSTM  & $k$-NN &    SVM &    MLP & 1D~CNN\\
        \hline
        4000~ms &       96 & 93.5\% & 97.9\% &100.0\% & 78.4\% &100.0\%\\
        2000~ms &       96 & 95.3\% & 98.7\% &100.0\% & 86.7\% & 99.7\%\\
        1000~ms &       96 & 98.4\% & 98.4\% &100.0\% & 89.6\% & 99.7\%\\
        500~ms  &       96 & 96.6\% & 98.2\% &100.0\% & 98.7\% &    n/a\\
        1~ms    &       96 & 88.5\% & 97.1\% &100.0\% & 99.0\% &    n/a\\
        \hline
        4000~ms &       64 & 90.9\% & 97.9\% &100.0\% & 74.0\% & 98.7\%\\
        2000~ms &       64 & 76.3\% & 99.0\% & 99.5\% & 78.1\% & 98.4\%\\
        1000~ms &       64 & 89.8\% & 98.2\% & 99.7\% & 85.9\% & 99.0\%\\
        500~ms  &       64 & 91.9\% & 98.2\% & 99.7\% & 96.4\% &    n/a\\
        1~ms    &       64 & 82.3\% & 97.9\% & 99.7\% & 97.4\% &    n/a\\
        \hline
        4000~ms &       32 & 80.2\% & 96.9\% & 99.0\% & 58.3\% & 94.8\%\\
        2000~ms &       32 & 73.7\% & 96.6\% & 99.0\% & 81.5\% & 95.8\%\\
        1000~ms &       32 & 81.5\% & 96.6\% & 99.5\% & 88.5\% & 96.4\%\\
        500~ms  &       32 & 76.3\% & 96.9\% & 99.5\% & 93.0\% &    n/a\\
        1~ms    &       32 & 66.4\% & 96.6\% & 99.2\% & 89.3\% &    n/a\\
        \hline
        4000~ms &       24 & 84.4\% & 95.8\% &100.0\% & 64.8\% & 95.3\%\\
        2000~ms &       24 & 84.4\% & 96.4\% &100.0\% & 78.4\% & 94.0\%\\
        1000~ms &       24 & 72.9\% & 95.1\% & 99.7\% & 83.6\% & 90.6\%\\
        500~ms  &       24 & 84.6\% & 96.6\% &100.0\% & 95.3\% &    n/a\\
        1~ms    &       24 & 60.9\% & 97.9\% & 99.2\% & 88.5\% &    n/a\\
        \hline
        4000~ms &       16 & 68.8\% & 93.8\% & 99.0\% & 66.9\% & 91.4\%\\
        2000~ms &       16 & 73.7\% & 93.5\% & 99.7\% & 83.1\% & 89.1\%\\
        1000~ms &       16 & 66.9\% & 93.5\% & 98.2\% & 91.7\% & 90.4\%\\
        500~ms  &       16 & 68.0\% & 93.2\% & 99.0\% & 88.8\% &    n/a\\
        1~ms    &       16 & 54.4\% & 94.5\% & 96.1\% & 84.9\% &    n/a\\
        \hline
        4000~ms &        8 & 68.8\% & 94.8\% & 99.0\% & 75.5\% & 90.9\%\\
        2000~ms &        8 & 57.3\% & 92.4\% & 97.9\% & 81.3\% & 87.5\%\\
        1000~ms &        8 & 58.9\% & 93.0\% & 98.2\% & 89.8\% & 88.8\%\\
        500~ms  &        8 & 63.5\% & 93.5\% & 98.2\% & 89.6\% &    n/a\\
        1~ms    &        8 & 52.1\% & 93.2\% & 92.4\% & 80.7\% &    n/a\\
      \end{tabular}
  \end{tabular}}
\end{table*}

Third, we reran the code released by \citet{spampinato2017} (an LSTM combined
with a fully connected layer and a ReLU layer) on the data released by
\citet{spampinato2017} but first applied various highpass filters with 14~Hz,
10~Hz, and 5~Hz cutoffs to the data.
Recall, from Table~\ref{tab:classifiers}, that we obtain a classification
accuracy of 93\% without such highpass filtering.
With the highpass filtering, classification accuracy drops to 32.4\%
(14~Hz), 29.8\% (10~Hz), and 29.7\% (5~Hz).

\subsection{Regression}
\label{sec:regression}

In support of claim~\ref{enum:b}, \citet[\S~3.3 and \S~4.2]{spampinato2017}
report an analysis whereby they use the LSTM, combined with a fully connected
layer and a ReLU layer, that was trained on EEG data as an encoder to produce a
128-element encoding vector for each image in their dataset.
They then regress the 1,000-element output representation from a number of
existing deep-learning object classifiers that have been pretrained on ImageNet
to produce the same encoding vectors.
When training this regressor, in some instances, they freeze the parameters of
the existing deep-learning object classifiers, while in other instances they
fine tune them while learning the regressor.
They report a mean square error (MSE) between 0.62 and 7.63 on the test set
depending on the particulars of the model and training regimen
\citep[Table~4]{spampinato2017}.
They claim that this result supports the conclusion that this is the
\emph{the first human brain--driven automated visual classification method}
and thus enables \emph{automated visual classification in a ``brain-based
  visual object manifold''} \citep[\S~5 p.~6816]{spampinato2017}.

\begin{figure*}
  \centering
  \includegraphics[width=\linewidth]{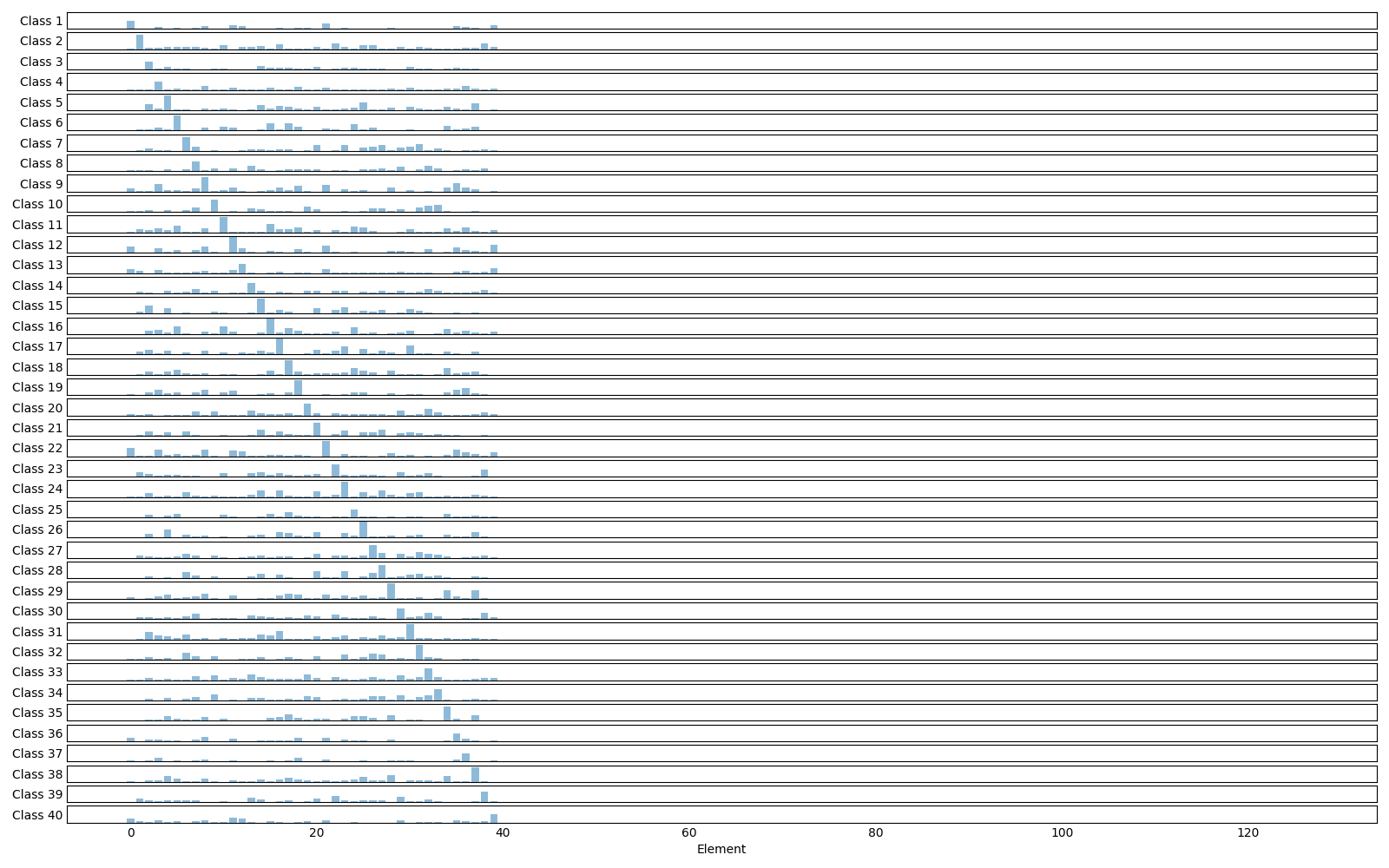}
  \caption{Encodings of the measured EEG response to each of the 40~ImageNet
    classes, as produced by the LSTM-based encoder of \citet{spampinato2017} on
    the EEG data released by \citet{spampinato2017}, averaged across all six
    subjects and all 50 images per class.}
  \label{fig:encodings}
\end{figure*}

Note that \citet{spampinato2017} use the same LSTM combined with a fully
connected layer and a ReLU layer both as a classifier and as an encoder.
During training as a classifier, the output of the last layer of the
classifier, namely the ReLU, is trained to match the class label.
Thus using such a trained classifier as an encoder would tend to encode EEG
data in a representation that is close to class labels.
Crucially, the output of the classifier taken as an encoder contains mostly, if
not exclusively, class information and little or no reflection of other
non-class-related visual information.
Further, since the output of their classifier is a 128-element vector, since
they have 40~classes, and since they train with a cross-entropy loss that
combines log softmax with a negative log likelihood loss, the classifier tends
to produce an output representation whose first 40 elements contain an
approximately one-hot-encoded representation of the class label, leaving the
remaining elements at zero.
Indeed, we observe this property of the encodings produced by the code released
by \citet{spampinato2017} on the data released by \citet{spampinato2017}
(Fig.~\ref{fig:encodings}).
Note that the diagonal nature of Fig.~\ref{fig:encodings} reflects an
approximate one-hot class encoding.
Any use of a classifier trained in this fashion as an encoder would have this
property.
\citet[\S~3.3, \S~4.2, and \S~4.4]{spampinato2017} use such an encoder to train
an object classifier with EEG data, \citet{palazzo2017}, \citet{kavasidis2017},
and \citet{tirupattur2018} use such an encoder to train a variational
autoencoder (VAE) \citep{kingma2013} or a generative adversarial network (GAN)
\citep{goodfellow2014} to produce images of human perception and thought, and
\citet{palazzo2018} use such an encoder to produce saliency maps, EEG
activation maps, and to measure association between EEG activity and layers in
an object detector.
Thus all this work is essentially driven by encodings of class information that
lack any visual information or any representation of brain processing.

We ask whether there is merit in the regression algorithm proposed by
\citet{spampinato2017} to create a novel object classifier driven by brain
signals.
We analyze their algorithm under the assumption that it is applied to EEG data
that supports classification of visually perceived objects and does not suffer
from contamination.
Under this assumption, the EEG response of two images of the same class would
be closer than for two images of different classes.
An encoder like the one employed by \citet{spampinato2017} would produce
encodings that are more similar for images of the same class than images of
different classes.
(For their actual encoder, Fig.~\ref{fig:encodings} shows that they are
indeed little more than class encodings).
Moreover, deep-learning object classifiers presumably produce closer
representations for images in the same object class than for images of
different classes.
After all, that is what object classifiers do.
Thus all the regressor does is preserve the property that two images of the
same class regress to closer representations than two images of different
classes.
In other words, all the regressor does is map a 1,000-dimension representation
of class to a 128-dimension representation of class.
It should not matter whether the actual target representation is a reflection
of brain processing or not.

We asked whether the putative success of this regression analysis depended on a
representation derived from neuroimaging.
To this end, we generated a random codebook with random codewords that simulate
the EEG response of all six subjects to all 2,000 image stimuli.
This was done with the following procedure.
We first generated 40 random codewords, one for each class, by uniformly
sampling elements i.i.d in $[0,2]$.
We then generated $\textrm{50}\times\textrm{6}=\textrm{300}$ random codewords
for each class, one for each subject and image, by adding univariate Gaussian
noise with $\sigma^2=4$ i.i.d. to the elements of the class codewords, and
clipped the elements to be nonnegative.
This generated a codebook of 12,000 random codewords for each simulated subject
response that has the property that encodings for images in the same class are
closer than entries for images in different classes.
These codewords carry no brain-inspired meaning whatsoever.
Like \citet{spampinato2017}, we then averaged the codewords across subject for
each image.
We then applied the PyTorch VGG-16 \citep{simonyan2014} pretrained on
ImageNet, without any fine tuning, to each of the images in the
\citet{spampinato2017} dataset.
Finally, we trained a linear regressor with MSE loss and L2 regularization from
the output of VGG-16 on each image to the average random codeword for that
image on the training set for the first split provided by
\citet{spampinato2017}.
We then measured an average MSE of 0.55 on the validation and test sets of that
split.
The fact that it is possible to regress the output of an off-the-shelf
pretrained object classifier to random class encodings as well as one can
regress that output to class encodings derived from an EEG encoder demonstrates
that the ability to do so does not depend on anything other than class
information in the source and target representations.

\subsection{Transfer learning}
\label{sec:transfer}

In further support of claim~\ref{enum:b}, \citet[\S~4.4]{spampinato2017} report
an analysis that purports to demonstrate that the learned combination of
regressor and object classifier generalizes to other datasets with disjoint
sets of classes.
To this end, they first apply VGG-16, pretrained on ImageNet, to a subset of the
Caltech~101 \citep{feifei2006} dataset with 30~classes, not fine tuned, to
produce a 1,000-element representation of each image.
They then map this with their regressor trained as described above to
128-element encodings.
Finally, they train and test an SVM classifier on the resulting encodings.
They compare this with an SVM classifier trained and tested on the 1,000-element
outputs from pretrained deep-learning object classifiers that have not been
mapped with their regressor and achieve comparable performance (92.6\% on the
1,000-element output of GoogleNet and 89.7\% on the 128-element encodings
regressed from GoogleNet).
They claim that their approach enables \emph{automated visual classification in
  a ``brain-based visual object manifold''} and \emph{show[s] competitive
  performance, especially as concerns learning EEG representation of object
  classes} \citep[\S~5 p.~6816]{spampinato2017}.

We conjecture that the putative success of this transfer-learning analysis is
not surprising and demonstrates nothing about the quality of the representation
nor whether it reflects brain processing.
As discussed above, the deep-learning object classifiers produce closer output
representations for images in the same object class than for images of
different classes.
Further, as discussed above, all the regressor does is preserve the property
that two images of the same class regress to closer encodings than two images
of different classes.
The choice of regressor or regressed representation should have no impact on
the SVM classifier so long as these properties hold.

We thus asked whether the putative success of this transfer-learning analysis
depended on a representation derived from neuroimaging.
To this end, we used VGG-16, pretrained on ImageNet without any fine tuning,
to map the images in Caltech~101 to 1,000-element encodings and applied the
regressor that we trained on random representations to map these 1000-element
encodings to 128-element encodings.
This composite mapping exhibited the above properties.
This, again, generated a codebook of random codewords for each image in this
subset of Caltech~101 that has the property that entries for images in the
same class are closer than entries for images in different classes.
As before, the codewords carry no brain-inspired meaning.
We split our subset of Caltech~101 into disjoint training and test sets,
trained a linear SVM on the training set, and achieved an accuracy of 95.9\% on
the test set when classified on the 128-element encodings regressed from VGG-16
as compared with 94.9\% on the test set when classified on the 1,000-element
output of VGG-16.

\section{Reconciling Discrepancies}
\label{sec:reconciling}

A number of papers, \eg\ \citet[Figs.~1 and~2(c)]{spampinato2016},
\citet[Figs.~1 and~2]{palazzo2017}, and \citet[Figs.~2, 3,
  and~4]{kavasidis2017}, use an encoder that appears to be similar or identical
to that reported in \citet[Figs.~2
  and~3(c)]{spampinato2017}.\footnote{\citet{palazzo2017} and
  \citet{tirupattur2018} appear to employ related but somewhat different
  encoders.
  We do not comment on these here since we do not have access to this code.}
A number of papers \citep{palazzo2017, kavasidis2017, du2018, palazzo2018} use
the dataset reported in \citet{spampinato2017}.\footnote{An earlier but similar
  dataset was reported in \citet{spampinato2016}.
  \citet{tirupattur2018} use a different dataset reported by \citet{kumar2018}.
  We do not comment on these here since we do not have access to these
  datasets.}
\citet{spampinato2017} have released the code\footref{foot:code} for their
encoder as well as their data\footref{foot:data}.
They have released their data in two formats, Python and Matlab.
We have observed a number of discrepancies between the different published
accounts, between the different released variants of the data, and between the
published accounts and the released code and data.
We discuss here how we reconciled such for the purposes of the experiments and
analyses reported here.
We do this solely to document precisely what we have done.
We do not believe that anything substantive turns on these issues, except for
the issue of filtering, whether or not the DC and VLF components are removed
from the EEG data.
In this case, we perform all analyses twice, with and without such removal.

\subsection{Filtering}
\label{sec:filtering}

\citet[\S~3.1 p.~7]{spampinato2016} state:
\begin{quote}
  \emph{A notch filter (49--51~Hz) and a second-order band-pass Butterworth
    filter (low cut-off frequency 14~Hz, high cut-off frequency 71~Hz) were set
    up so that the recorded signal included the Beta (15--31~Hz) and Gamma
    (32--70~Hz) bands, as they convey information about the cognitive processes
    involved in the visual perception [15].}
\end{quote}
\citet[\S~3.1 p.~6812]{spampinato2017} state:
\begin{quote}
  \emph{A notch filter (49--51~Hz) and a second-order band-pass Butterworth
    filter (low cut-off frequency 14~Hz, high cut-off frequency 71~Hz) were set
    up so that the recorded signal included the Beta (15--31~Hz) and Gamma
    (32--70~Hz) bands, as they convey information about the cognitive processes
    involved in the visual perception [15].}
\end{quote}
\citet[\S3.1 p.~3412]{palazzo2017} state:
\begin{quote}
  \emph{The acquired EEG signals were filtered in run-time (\ie\ during the
    acquisition phase) by the integrated hardware notch filter (49--51~Hz) and a
    second order Butterworth (band-pass) filter with frequency boundaries
    14--70~Hz.
    This frequency range contains the necessary bands (Alpha, Beta and Gamma)
    that are most meaningful during the visual recognition task [17].}
\end{quote}
Later publications \citep{kavasidis2017, palazzo2018, tirupattur2018} do not
discuss filtering.

\begin{figure*}
  \centering
  \begin{tabular}{@{}cccccc@{}}
    \includegraphics[width=0.15\linewidth]{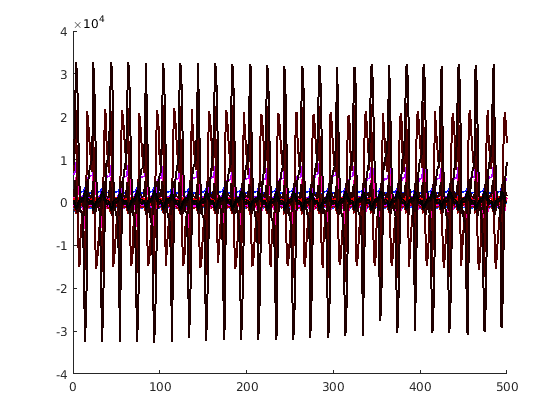}&
    \includegraphics[width=0.15\linewidth]{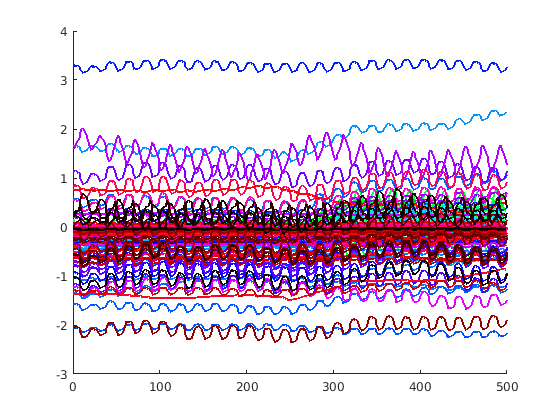}&
    \includegraphics[width=0.15\linewidth]{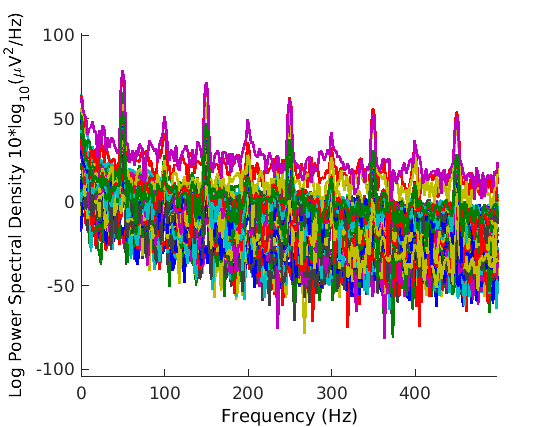}&
    \includegraphics[width=0.15\linewidth]{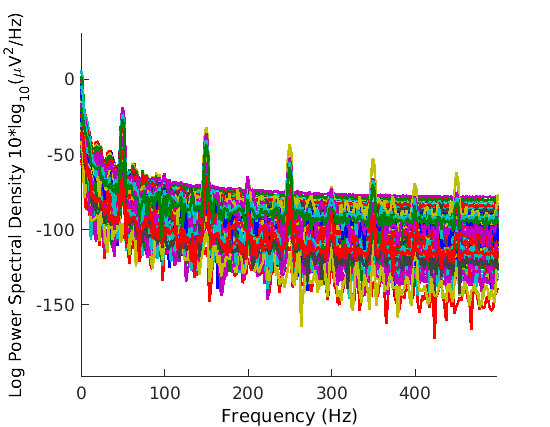}&
    \includegraphics[width=0.15\linewidth]{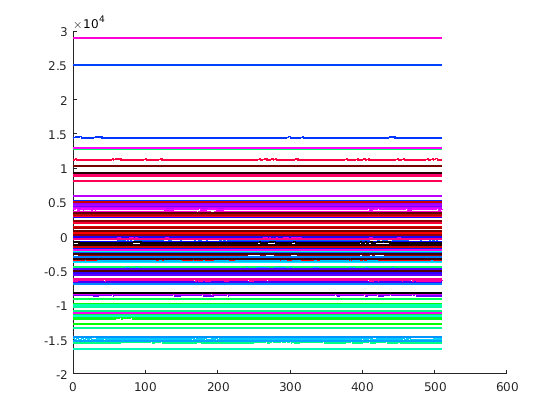}&
    \includegraphics[width=0.15\linewidth]{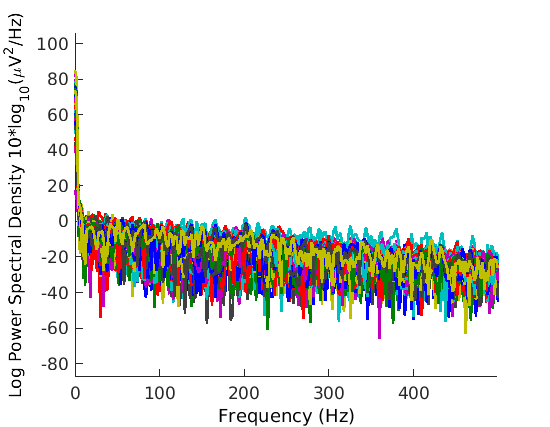}\\
    (a)&(b)&(c)&(d)&(e)&(f)\\
  \end{tabular}
  \caption{Waveforms of the EEG data for the ImageNet image stimulus
    \texttt{n02124075\_12647}, as provided by \citet{spampinato2017} for
    Subject~1, in (a)~Python format and (b)~Matlab format.
    Corresponding spectra of the (c)~Python format and (d)~Matlab format data.
    (e)~Waveforms and (f)~spectra for the same stimulus in our raw unfiltered
    data collected with a block design.}
  \label{fig:spectra}
\end{figure*}

A spectral analysis (Fig.~\ref{fig:spectra}) of both the Python and Matlab
format data released by \citet{spampinato2017} suggests that neither have been
filtered by any notch filter nor any bandpass filter.
The Python and Matlab format data differ greatly, and have somewhat different
spectra.
The code originally released by \citet{spampinato2017} does not contain any
notch filtering or any bandpass filtering, but does contain z-scoring.
We asked the authors to clarify \citep{kavasidis2018a}.
Email reply from \citet{spampinato2018a} stated:
\begin{quote}
  \texttt{We did very little pre-processing (notch filtering and normalization)
    on the data.}
\end{quote}
We provided \citep{spampinato2018e} an early draft of this manuscript to the
authors of \citet{spampinato2017}.
That draft pointed out the spectral analysis.
Email response from \citet{palazzo2018a} stated:
\begin{quote}
  \texttt{Our released dataset is not preprocessed but raw, directly coming
  from the EEG device that does not allow us to perform the erroneous
  filtering you are suggesting.}
\end{quote}
In further response, email from \citet{spampinato2018b} stated:
\begin{quote}
  \texttt{As we said to you in our previous email exchanges, we did not perform
    any pre-processing except notch filtering and normalization (there is a
    misprint in our original paper, despite in the work webpage we haven't
    reported that filtering) and this processing was done during models'
    training on the raw EEG data (the one available online).}
\end{quote}
In response to further requests for clarification, email from
\citet{spampinato2018c} stated:
\begin{quote}
  \texttt{all the results reported in the paper are obtained with the code
    http://perceive.dieei.unict.it/files/
    cvpr\_2017\_eeg\_encoder.py (with the
    exception that we performed notch filtering with an additional code that it
    is not online but we can send to you) on the .pth data at
    https://cloud.perceive.dieei.unict.it/
    index.php/s/XFF23FHapphsTut/download.
    Results may be slightly different because different hyperparameters and
    epochs.}

  \texttt{We haven't done any analysis on the matlab file (I suggest you not to
    use it). We also noticed that that data is slightly different (despite the
    dynamic is the same) from the python version (but we only opened the raw
    file with EEGLab and export it as mat file).}

  \texttt{No filtering was applied at the acquisition time and before
    generating the python data available online.}
\end{quote}
Further email from \citet{spampinato2018d} stated:
\begin{quote}
  \texttt{attached you can find the revised version of the
    http://perceive.dieei.unict.it/files/
    cvpr\_2017\_eeg\_encoder.py which
    includes notch filtering and [0-71]hz passband filtering that you can use
    for your tests.
    This newest code is written in PyTorch 0.4 (the online version is 0.3)
    which should be more efficient.}
\end{quote}
We take this all to imply that no filtering was applied during acquisition,
no filtering was applied prior to production of either the Python or Matlab
format released data, the analyses reported in \citet{spampinato2017} were
performed using the released code version 0.3 which did not perform any
filtering, and any filtering code was added subsequent to our contact with
\citet{spampinato2017}.

Examination of version 0.4 of the code provided to us indicates that the data
was first z-scored, then notch filtered (47--53~Hz), and then lowpass filtered
(71~Hz), all on a segment-by-segment basis.
This is unconventional in five ways.
First, z-scoring would normally be done after filtering, not before.
Second, z-scoring would normally be done after segmentation, not before.
Third, filtering would normally be done before segmentation, not after.
Fourth, the notch filter employed a different notch band than previously
reported.
Fifth, the data was lowpass filtered rather than bandpass filtered.
All analyses reported here were performed with the released (version 0.3)
code, modified as discussed below, on the Python format data, unmodified,
except as discussed below and in the text.

\subsection{Quantization}

\citet[\S~3.1 p.~6813]{spampinato2017} state:
\begin{quote}
  \emph{Data value distribution was centered around zero, thus
    non-linear quantization was applied.}
\end{quote}
\citet[\S~3.1 p.~3413]{palazzo2017} similarly state:
\begin{quote}
  \emph{The histogram of the acquired signals over the different
    values presented with a high density near the zero value and a much lower
    density at the extremities. In order to reduce input space sparsity,
    non-uniform quantization was applied for data compression.}
\end{quote}
As the released code contains no indication of such, we have no way of knowing
sufficient details of how to replicate this quantization on our data.
We further have no way of knowing if the released Python and/or Matlab data
reflects this quantization or not.
Thus we do not perform any quantization on either the released data or our new
data as part of any analyses reported here.

\subsection{Trials considered}

\citet{spampinato2017} nominally collected 50 trials for each of 40 stimuli
and 6 subjects for a total of 12,000 trials.
However, \citet[\S~3.1 p.~3413]{palazzo2017} state:
\begin{quote}
  \emph{Given that the systems involved are not real-time (Operating system
    process scheduler, DAQ hardware etc\ldots), variable length EEG sequences
    were dealt with by discarding those with less than 480 samples.
    Data sequences whose length was between 480 and 500 samples w[h]ere padded
    with zeros until reaching 500 samples. Sequences longer than 500 samples
    were tail trimmed.
    [\ldots]
    534 samples did not satisfy the minimum data length
    criteria described above, resulting in 11,466 valid
    samples.}
\end{quote}
\citet[\S~3.1 p.~1811]{kavasidis2017} further state:
\begin{quote}
  \emph{After the EEG data acquisition, we obtained 11,466 128-channel EEG
    sequences (536 recordings were discarded because they were too short or too
    altered to be included in the experiment).}
  \citep[\S~3.1 p.~3413]{palazzo2017}
\end{quote}
The released data in Python format contains 11,965 trials which is a
superset of the released data in Matlab format that contains 11,466 trials.
Further, the Python format data differs from the Matlab format data.
We have no way of knowing why the data differs, why the Python format data
contains 11,965 trials, and whether the 11,466 trials in the Matlab format data
correspond to the 11,466 trials reported above.
Nonetheless, we take 536 \vs\ 534 to be a typo and use the Python format data,
but only those 11,466 trials that appear in the Matlab format data.
The 499 trials thus discarded come from Subject~2.

\subsection{Trial window}

The published papers indicate that samples 40--480 were used.
\citet[\S~3.1 p.~6813]{spampinato2017} state:
\begin{quote}
  \emph{From each recorded EEG sequence, the first 40 samples (40 ms) for
    each image were discarded in order to exclude any possible interference from
    the previously shown image (\ie\ to permit the stimulus to propagate from
    the retina through the optical tract to the primary visual cortex [8]).
    The following 440 samples (440 ms) were used for the experiments.}
\end{quote}
\citet[\S~3.1 p.~3413]{palazzo2017} further state:
\begin{quote}
  \emph{From each recorded EEG sequence, the first 40 samples were discarded
    in order to minimize any possible interference from the previously shown
    image (\ie\ to give the necessary time for the stimulus to clear its way
    through the optical tract [9]).
    The following 440 samples (440 ms) were used for the experiments.}
\end{quote}
\citet[\S7.1 p.~7]{palazzo2018} further state:
\begin{quote}
  \emph{The exact duration of each signal may vary, so we discard the first 40
    samples (40 ms) to reduce interferences from the previous image and then
    cut the signal to a common length of 440 samples (to account for signals
    with $L < \text{500}$), when supposedly all image-related visual and
    cognitive processes will have been completed.}
\end{quote}
The released code, however, uses samples 20--450 (\ie\ a sequence of length
430), lacks zero padding and tail trimming, and discards sequences shorter than
450 samples or longer than 600 samples.
No trials are shorter than 480 samples so none are are discarded for this
reason and none require zero padding.
The released code, however, discards 25 trials beyond the 534 mentioned
above for being longer than 600 samples.
We have no way of knowing what was actually done to obtain the results in
\citet{spampinato2017}, \citet{palazzo2017}, \citet{kavasidis2017}, and
\citet{palazzo2018}.
Here, we modified the released code to not discard (the 25) trials longer
than 600 samples and to use samples 40--480 from each trial instead of
20--450.

\subsection{The encoder model}

When describing the encoder model (\citep[Figs.~1 and~2(c)]{spampinato2016},
\citep[Figs.~1 and~2]{palazzo2017}, and \citep[Figs.~2, 3,
  and~4]{kavasidis2017}), \citet[\S~3.2 p.~9]{spampinato2016} state:
\begin{quote}
  \emph{an additional output layer (linear combinations of input, followed by
  ReLU nonlinearity) is added after the LSTM}
\end{quote}
\citet[\S~3.2 p.~6813]{spampinato2017} similarly state:
\begin{quote}
  \emph{an additional output layer (linear combinations of input, followed by
  ReLU nonlinearity) is added after the LSTM}
\end{quote}
\citet[\S~3.2 p.~3414]{palazzo2017} similarly state:
\begin{quote}
  \emph{the final output state of the LSTM goes into a fully-connected layer
  with ReLU non-linearity.}
\end{quote}
\citet[\S~3.2 p.~1811]{kavasidis2017} similarly state:
\begin{quote}
  \emph{The first processing module of our approach consists of an encoder,
  which receives as input an EEG time series and provides as output a more
  compact and class-discriminative feature vector.
  In [26] we tested several encoder models and the most performing one is
  shown in Fig. 4.
  It consists of a standard LSTM layer followed by a nonlinear layer.
  An input EEG sequence is fed into the LSTM layer, whose output at the final
  time step goes into a fully-connected layer with a ReLU activation function.
  This simple architecture when stacked with a 40-way softmax layer yielded
  good performance---over 80\% classification accuracy.}
\end{quote}
However, the released code omits the ReLU layer.
We modified the released code to add the ReLU layer for the analyses reported
here.

\subsection{The classifier}

\citet[Fig.~1]{spampinato2016} and \citet[Fig.~2]{spampinato2017} report
training the encoder by attaching a classifier to its output and training
against known labels.
\citet[\S~3.2 p.~8]{spampinato2016} state:
\begin{quote}
  \emph{The encoder network is trained by adding, at its output, a
  classification module (in all our experiments, it will be a softmax layer),
  and using gradient descent to learn the whole model/s parameters end-to-end.}
\end{quote}
\citet[\S~3.3 p.~10]{spampinato2016} state:
\begin{quote}
  \emph{modified it by replacing the softmax classification layer with a
  regression layer}
\end{quote}
\citet[\S~4.3 p.~14]{spampinato2016} state:
\begin{quote}
  \emph{Our automated visual classifier consists of the combination of the
  CNN-based feature regressor achieving the lowest MSE (GoogleNet features
  with k-NN regressor, trained on average features) with the softmax
  classifier trained during EEG manifold learning.}
\end{quote}
\citet[\S~3.2 p.~6813]{spampinato2017} state:
\begin{quote}
  \emph{The encoder network is trained by adding, at its output, a
    classification module (in all our experiments, it will be a
    softmax layer), and using gradient descent to learn the whole model's
    parameters end-to-end.}
\end{quote}
\citet[\S~3.3 p.~6814]{spampinato2017} state:
\begin{quote}
  \emph{modified it by replacing the softmax classification layer with a
    regression layer}
\end{quote}
\citet[\S~3.3 p.~6814]{spampinato2017} state:
\begin{quote}
  \emph{The resulting CNN-based regressor is able to extract brain-learned
    features from any input image for fu[r]ther classification by the softmax
    layer trained during EEG feature learning.}
\end{quote}
\citet[\S~4.3 p.~6816]{spampinato2017} state:
\begin{quote}
  \emph{and then classifies feature vectors using the softmax classifier
    trained during EEG manifold learning.}
\end{quote}
\citet[\S~3.2 p.~3414]{palazzo2017} state:
\begin{quote}
  \emph{We append a softmax classification layer and perform gradient descent
    optimization (supervised by the class of the image shown when
    the input signal had been recorded) to train the encoder and the classifier
    end-to-end.}
\end{quote}
\citet[\S~3.2 p.~1811]{kavasidis2017} state:
\begin{quote}
  \emph{This simple architecture when stacked with a 40-way softmax
    layer yielded good performance --- over 80\% classification accuracy.}
\end{quote}
\citet[\S~4.3 p.~1815]{kavasidis2017} state:
\begin{quote}
  \emph{with the softmax layer changed for a 40-class classification task.}
\end{quote}
\citet[\S~4.3 p.~954]{tirupattur2018} state:
\begin{quote}
  \emph{We use ReLU activation for all the layers in our network and Softmax
  for the final classification layer.}
\end{quote}
\citet[\S~7.2 p.~8]{palazzo2018} state:
\begin{quote}
  \emph{Once training is completed, we use the trained EEG and image encoders
  as feature extractors in the joint embedding space, followed by a softmax
  layer, for both image and EEG classification.}
\end{quote}
\citet[\S~7.2 p.~8]{palazzo2018} state:
\begin{quote}
  \emph{Both our model and pre-trained visual encoders are used as feature
  extractors followed by a softmax layer}
\end{quote}
The released code appears to use PyTorch
\texttt{torch.nn.functional.cross\_entropy}, which internally uses
\texttt{torch.nn.functional.log\_softmax}.
This is odd for two reasons.
First, this has no parameters and does not require any training.
Second, training a 40-way classifier this way, appended to an encoder, with an
implicit one-hot representation of class labels, will tend to train the encoder
to produce 128-element EEG encodings where all but the first 40 elements are
zero (Fig.~\ref{fig:encodings}).
Indeed, we have observed this behavior with the released code.
We have no way of knowing what was actually intended and used to generate the
reported results.
Here, like the released code, we train the encoders with the same cross-entropy
loss, which internally contains a log softmax operation, but use the output of
the encoder, prior to any softmax operation, for classification.
(Note that had the output of the softmax layer been taken as the EEG encodings,
they would have been one-hot.)

\section{Discussion}
\label{sec:discussion}

The analyses in \S~\ref{sec:reanalysis} demonstrate that the results reported
by \citet{spampinato2017} do not depend on the temporal structure of the EEG
signal.
The analyses in \S~\ref{sec:newAnalysis} demonstrate that the results reported
by \citet{spampinato2017} crucially depend on a block design and cannot be
replicated with a rapid-event design.
However, the block design of \citet{spampinato2017}, together with their
training/test-set splits, is such that every trial in each test set comes from
a block that has many trials in the corresponding training set.
The first analysis in \S~\ref{sec:design} shows that if one adopts splits that
separate trials from a block so that the test sets never contain trials from
blocks that have any trials in the corresponding training sets, classification
accuracy drops to chance.
This strongly suggests that the high classification accuracy obtained by
\citet{spampinato2017} crucially depends on such contamination, which
constituted surreptitious training on the test set.
This is further corroborated by the second analysis in \S~\ref{sec:design} that
shows that one can obtain near perfect classification accuracy with an
experiment design where labels vary only by block but where the class of the
stimuli within the block are uncorrelated with the labels.
If the methods of \citet{spampinato2017} were indeed classifying brain activity
due to perception of the class of the stimuli, one would expect to obtain
chance performance with this analysis.
The fact that near perfect performance was obtained strongly suggests that
these methods are indeed classifying the long-term static brain activity that
persists during a block that is uncorrelated with the perceptual activity.
Finally, the third analysis in \S~\ref{sec:design} shows that this finding is
exacerbated by the presence of DC and VLF components of the recorded EEG signal
that are present due to the omission of bandpass filtering.
We propose that all future classification experiments performed on EEG data
employ a design that controls for such contamination.
Since the data released by \citet{spampinato2017} irreparably suffers from this
contamination, it renders this dataset unsuitable for its intended purpose of
decoding perceptual and conceptual processing and further invalidates all
subsequent analyses and claims that use this data for those purposes
\citep{palazzo2017, kavasidis2017, du2018, palazzo2018}.

\subsection{Consequences of flawed filtering}

While \citet{spampinato2017} and two related papers \citep{spampinato2016,
  palazzo2017} suggest that the reported results were obtained with a process
that included notch and bandpass filtering, subsequent analysis and
communication with the authors now suggest that this was not the case
(\S~\ref{sec:filtering}).
This analysis and communication with the authors has led them to modify their
code (\S~\ref{sec:filtering}).
This is important for two reasons.
First, the modifications that they made do not address the issue at hand.
The added lowpass (71~Hz) filter does not address the data contamination problem
resulting from the block design that leads to surreptitious training on the
test set.
Further, it preserves the DC and VLF components that exacerbate that problem.
It is those components, not the HF components, that need to be removed.
Even doing this would only address the exacerbation.
It would not address the root cause which is the block design.
Second, the fact that the authors omitted the bandpass filter exacerbated the
issue, leading to egregious overestimation of the classification accuracy.
This has led to their results and data receiving considerable attention and
enthusiasm, possibly contributing to the sheer number of papers that use this
dataset and/or pursue similar approaches.
Had the stated filtering been performed, perhaps the resulting more modest (but
still invalid) results would have tempered the rapid proliferation of follow-up
work that also suffers from similar methodological shortcomings.

\subsection{Consequences of flawed block design on subsequent papers}

The above strongly suggests that the output of the LSTM-based encoder trained
by \citet{spampinato2017} does not constitute a \emph{``brain-based visual
  object manifold''} \citep[\S~5 p.~6816]{spampinato2017}.
Further, the analyses in \S~\ref{sec:regression} and~\ref{sec:transfer}
strongly suggest that the object classifiers constructed by \citet[\S~3.3,
  \S~4.2, and~\S~4.3]{spampinato2017} are not making use of \emph{any}
information in the output of the trained LSTM-based encoder, whether or not it
contains a representation of human brain processing.
Since these flaws are orthogonal to those of the data contamination issue,
these methods are irreparably flawed and their shortcomings would not be
remedied by correction of the contamination issue.

\citet{kumar2018} report a different EEG dataset that also appears to have been
collected with a block design.
Data was recorded from a single 10~s block for a single stimulus from each of
30~classes for each of 23~subjects.
Each 10~s block was divided into either 40 or 200 segments.
Ten-way cross validation was performed during analysis.
We have no way of knowing whether the test sets contained segments from the
same blocks that had segments in the corresponding training sets.
But since a single block was recorded for each stimulus for each subject, the
only way to avoid such would have been to conduct cross-subject analyses.
The first analysis in \S~\ref{sec:design} suggests that such cross-subject EEG
analysis is difficult and far beyond the current state of the art.

\citet{tirupattur2018} report using the dataset from \citet{kumar2018} to drive
a generative adversarial network (GAN) in a fashion similar to
\citet{palazzo2017}.
That work performs five-way cross validation during analysis.
Again, we have no way of knowing whether the test sets contained segments from
the same blocks that had segments in the corresponding training sets, and
avoiding such would have required cross-subject analyses that our experiments
suggest are far beyond the current state of the art.

\subsection{Consequences of using flawed EEG encodings as input to image
  synthesis}

\begin{figure*}
  \centering
  \resizebox{\linewidth}{!}{\begin{tabular}{@{}cc@{}}
    \includegraphics[width=\linewidth]{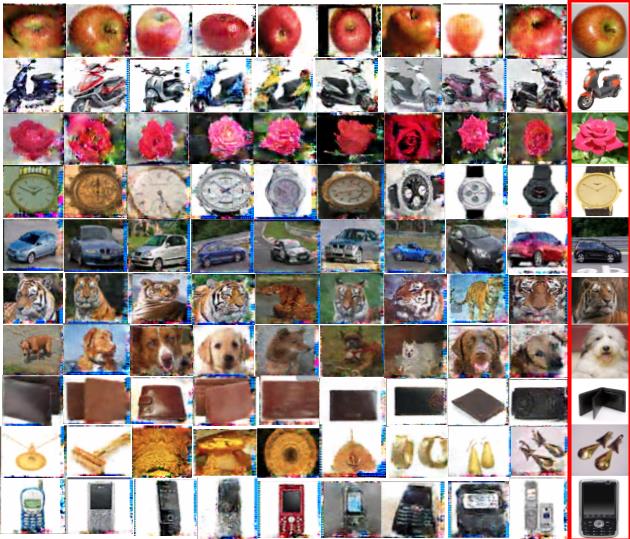}&
    \includegraphics[width=\linewidth]{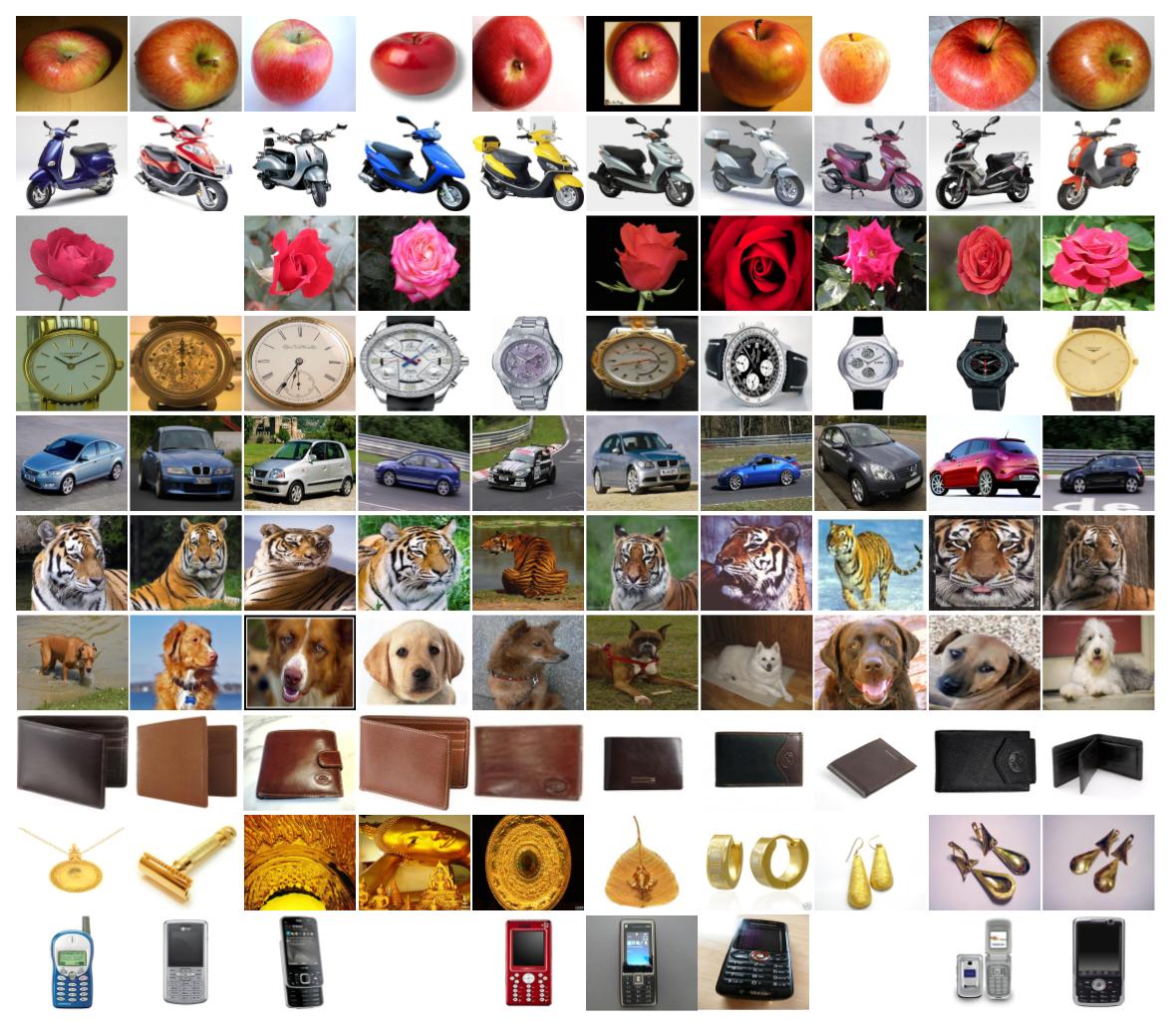}
  \end{tabular}}
  \caption{(left)~Fig.~6 from \citet{tirupattur2018} illustrating sample images
    purportedly generated by a GAN model from EEG encodings (except for the
    right column in red that illustrates a random image of the given class from
    the training data).
    (right)~corresponding identical ImageNet images for almost all of the
    generated images.
    Note that some, but not all, of the purportedly synthesized images on the
    left are horizontal mirrors of ImageNet images on the right.
    Also note that all of the purportedly synthesized images contain the same
    precise fine-grained detail as the corresponding ImageNet images.
    In particular, each image not only depicts the corresponding class but also
    depicts the exact non-class-specific background as the ImageNet
    counterpart.}
  \label{fig:gan}
\end{figure*}

\citet{palazzo2017}, \citet{kavasidis2017}, and \citet{tirupattur2018} all
purport to use the EEG encodings to generate images using a GAN that depict
human perception and thought.
Since we lack access to the code for any of these papers, we are unable to
perform the kind of random data analysis that we perform in
\S~\ref{sec:regression} and~\ref{sec:transfer} to evaluate these methods.
Instead, here we analyze the result in \citet{tirupattur2018}, using only the
published synthesized images.
We select this paper because it has the most extensive set of generated
examples.
\citet[abstract p.~950]{tirupattur2018} state:
\begin{quote}
  \emph{While extracting spatio-temporal cues from brain signals for
    classifying state of human mind is an explored path, decoding and
    visualizing brain states is new and futuristic.
    Following this latter direction, in this paper, we propose an approach that
    is able not only to read the mind, but also to decode and visualize human
    thoughts.
    More specifically, we analyze brain activity, recorded by
    an ElectroEncephaloGram (EEG), of a subject while thinking about
    a digit, character or an object and synthesize visually the thought
    item.
    To accomplish this, we leverage the recent progress of adversarial learning
    by devising a conditional Generative Adversarial
    Network (GAN), which takes, as input, encoded EEG signals and
    generates corresponding images.}
\end{quote}
\citet[\S~1 p.~950]{tirupattur2018} further state:
\begin{quote}
  \emph{Our goal is to extract some cues from the brain activity, recorded
    using low-cost EEG devices$^1$, and use them to visualize the thoughts of a
    person.
    More specifically, we attempt to visualize the thoughts of a person by
    generating an image of an object that the person is thinking about.
    EEG data of the person is captured while he is thinking of that
    object and is used for image generation.
    We use a publicly available
    EEG dataset [16] for our experiments and propose a generative
    adversarial model for image generation.
    We make the following contributions in this work: 1) we introduce the
    problem of interpreting and visualizing human thoughts, 2) we propose a
    novel conditional GAN architecture, which generates class-specific images
    according to specific brain activities; 3) finally, we also show that our
    proposed GAN architecture is well suited for small-sized datasets and can
    generate class-specific images even when trained on limited training data.}
  \par
  \emph{We demonstrate the feasibility and the effectiveness of the proposed
    method on three different object categories, i.e., digits, characters, and
    photo objects, and show that our proposed method is, indeed, capable of
    reading and visualizing human thoughts.}
\end{quote}
Conditional GANs are not intended to output exact copies of the training set
because the input that leads to synthesized images contains noise in addition to
class information.
GANs in their true spirit are supposed to learn visual features that are
indicative of different instances of objects within a class and synthesize
novel images for instances of a class by selecting and combining those
features in a semantically and visually coherent fashion.
The current state-of-the-art \citep{salimans2016} is unable to achieve this
lofty goal.
However, essentially all of the example images illustrated in
\citet[Fig.~6]{tirupattur2018} are nearly exact copies of images in ImageNet
(Fig.~\ref{fig:gan}).
GANs typically do not generate such near exact copies.
Moreover, in order for them to generate the same image twice, they must be
provided with the same conditioning input, which in this case comprises both an
EEG encoding and noise.
It would be highly unlikely for the same EEG encoding and the same noise to be
provided at each training iteration.
Thus it would be highly unlikely for a proper conditional GAN to be able to
memorize the training set.
Moreover, it would be highly unlikely for the same EEG encoding and the same
noise to be provided both during training and test.
Thus it would be highly unlikely for a proper conditional GAN to output near
exact copies of the training set during test.
Without their code and data, it is impossible for us to precisely determine the
cause of this highly unlikely circumstance.
Nonetheless, this calls into question their claim that their \emph{proposed
  method is, indeed, capable of reading and visualizing human thoughts}.

\subsection{Consequences for flawed joint training of EEG and image encoders to
  analyze brain processing of images}

\citet[Fig.~1]{palazzo2018} jointly train an EEG encoder and an image encoder
to produce similar encoded representations and then purport to use the trained
encoders for several purposes: producing saliency maps \citep[\S~4, \S~7.3, and
  Figs.~3 and~5]{palazzo2018}, producing EEG activation maps \citep[\S~5,
  \S~7.4, and Fig.~6]{palazzo2018}, and associating EEG activity with layers in
a synthetic object detector \citep[\S~6, \S~7.4, and Fig.~9]{palazzo2018}.
Since these results were all produced with the same contaminated dataset, these
results are all suspect.
Moreover, Tables~\ref{tab:newDataRapidEventBandpass}
and~\ref{tab:newDataRapidEventNoBandpass} suggest that using the proposed
methods to produce legitimate results from uncontaminated data collected with a
rapid-event design is unlikely to succeed.
Beyond this, however, the methods themselves appear to be fundamentally flawed
and unlikely to demonstrate anything, even if they could be made to work on
uncontaminated data.
The loss function employed in the joint training regimen simply constrains the
two encoded representations to be similar.
A perfectly trained image encoder, trained against class labels, would simply
encode image class, no more and no less.
A perfectly trained EEG encoder, trained against class labels, would simply
encode stimulus class, no more and no less.
During joint training of the EEG encoder, the image encoder serves simply as a
surrogate for class labels, no more and no less.
Similarly during joint training of the image encoder, the EEG encoder serves
simply as a surrogate for class labels, no more and no less.
Thus joint training accomplishes nothing that could not be accomplished by
training the components individually against class labels.
The resulting encoded representations  would contain no information beyond
class labels.
With this, the saliency map \citep[Eqs.~(3) and~(4) and Fig.~5]{palazzo2018}
measures nothing more than the degree to which image regions impact
classification accuracy of an object detector trained against class labels.
Brain activity, whether encoded in EEG data or not, plays no role in
constructing these saliency maps.
The importance~$I_c$ of an EEG channel~$c$ as rendered in activation maps
\citep[Eqs.~(5--7) and Fig.~6]{palazzo2018} measures nothing more than the
degree to which removing the information in~$c$ decreases the classification
accuracy, averaged over trials for a class and/or subjects.
While this nominally is a valid approach, with the contaminated data collected
with a block design, all these maps illustrate is the degree to which a given
channel encodes the arbitrary long-term brain states associated with the block,
not any class-specific information.
Moreover, Tables~\ref{tab:window}--\ref{tab:newDataBlockBandpass},
\ref{tab:newDataBlockNoBandpass},
and~\ref{tab:reanalysisA}--\ref{tab:reanalysisBnoBandpass}, suggest that any
purported temporal information in \citet[Figs.~7 and~9]{palazzo2018} is
artifactual.
Tables~\ref{tab:newDataRapidEventBandpass}
and~\ref{tab:newDataRapidEventNoBandpass} suggest that activation maps computed
with uncontaminated data collected with a rapid-event design would simply be
blank, as accuracy would be at chance levels both prior and subsequent to
removing the information in any particular EEG channel.
Finally, association~$A_{c,l}$ between an EEG channel~$c$ and any component~$l$
of an object detector is simply a linear combination of the class-average
activation maps \citep[Fig.~6]{palazzo2018} weighted by the degree to which
removing the portion~$l$ of an object detector causes misclassifications to a
given class.
This holds whether~$l$ is a portion of a feature map, an entire
feature map, or all feature maps in a given layer, as computed by
\citet[Eqs.~(8--10)]{palazzo2018} and rendered in \citet[Fig.~9]{palazzo2018}.
The fact that the activation maps in \citet[Fig.~9]{palazzo2018} become more
diffuse for later layers in an object detector says nothing more than the fact
that removing later layers in an object detector leads to higher entropy in the
output distribution, a property solely due to the image classifier and
completely independent of any brain processing, whether measured by EEG or not.

\subsection{Summary}

In summary, our results call into question not only the results of
\citet{spampinato2017} but other published results as well
\citep{spampinato2016, palazzo2017, kavasidis2017, du2018, kumar2018,
  tirupattur2018, palazzo2018}.
They do so in four distinct ways.
First, they raise doubts about all claims that depend directly or indirectly
on the ability to use the kinds of classification algorithms reported here,
including the particular classification algorithm of \citet{spampinato2017}, to
extract class information from the particular data of \citet{spampinato2017}.
That alone raises doubts about all of the above cited papers.
Second, they raise doubts about the ability of the kinds of classification
algorithms reported here, including the particular classification algorithm of
\citet{spampinato2017}, to extract class information from any EEG data
collected with a block design.
It places the burden of proof that there is no data contamination on any use of
a block design.
This raises doubts not just about the particular dataset collected by
\citet{spampinato2017}, but further about the experimental protocol proposed by
\citet{spampinato2017}.
Third, they demonstrate that a whole spectrum of classification algorithms do
not work on a dataset collected with a rapid-event design that does not suffer
from data contamination.
This raises doubts about not just the dataset and protocol, but further about
the analysis methods and algorithms.
Fourth, \S~\ref{sec:regression} and \S~\ref{sec:transfer} raise doubts about
the general approach underlying the proposed methods and algorithms for using
EEG data to advance computer vision.
While we have employed the random-data attack to the particular methods of
\citet{spampinato2017}, we believe that they also can be applied to all of the
methods in \citet{palazzo2017}, \citet{kavasidis2017}, \citet{tirupattur2018},
and \citet{palazzo2018} as well.
We are hindered in our attempt to conduct this analysis by the fact that the
authors have declined to release their code to us, despite requests, and the
fact that the published papers lack sufficient detail to replicate their
models.

\section{Conclusion}
\label{sec:conclusion}

The results in Tables~\ref{tab:newDataRapidEventBandpass}
and~\ref{tab:newDataRapidEventNoBandpass} suggest that the ability to classify
40~object classes in image stimuli and 12~activity classes in video stimuli
from an EEG signal is extremely difficult and well beyond the current state of
the art.
Moreover, the enterprise of using neuroimaging data to train better
computer-vision systems, proposed by \citep[\S~8 p.~625]{barbu2014} and
\citep[Fig.~2 and \S~3 last~\P p.~4068]{siskind2015}, requires more
sophisticated methods than simply attaching a regressor to a pretrained object
classifier and is also likely to be difficult and beyond the current state of
the art.
Both of these enterprises are the subject of substantial ongoing effort.
When widely published \citep{spampinato2016, spampinato2017, palazzo2017,
  kavasidis2017, du2018, kumar2018, tirupattur2018, palazzo2018}, inordinately
optimistic claims can lead to misallocation of valuable resources and can
sideline more modest but legitimate and important advances in the field.
Thus, when the sensational claims are recognized as false, it is imperative
that the refutation be widely publicized to appropriately caution the
community.

\ifCLASSOPTIONcompsoc
  \section*{Acknowledgments}
\else
  \section*{Acknowledgment}
\fi

This work was supported, in part, by the US National Science Foundation under
Grants 1522954-IIS and 1734938-IIS, by the Intelligence Advanced Research
Projects Activity (IARPA) via Department of Interior/Interior Business Center
(DOI/IBC) contract number D17PC00341, and by Siemens Corporation, Corporate
Technology.
Any opinions, findings, views, and conclusions or recommendations expressed in
this material are those of the authors and do not necessarily reflect the
views, official policies, or endorsements, either expressed or implied, of the
sponsors.
The U.S. Government is authorized to reproduce and distribute reprints for
Government purposes, notwithstanding any copyright notation herein.

\ifCLASSOPTIONcaptionsoff
  \newpage
\fi

\bibliographystyle{abbrvnat}
\bibliography{tpami2019}

\begin{IEEEbiography}
  [{\includegraphics[width=1in,height=1.25in,clip,keepaspectratio]
      {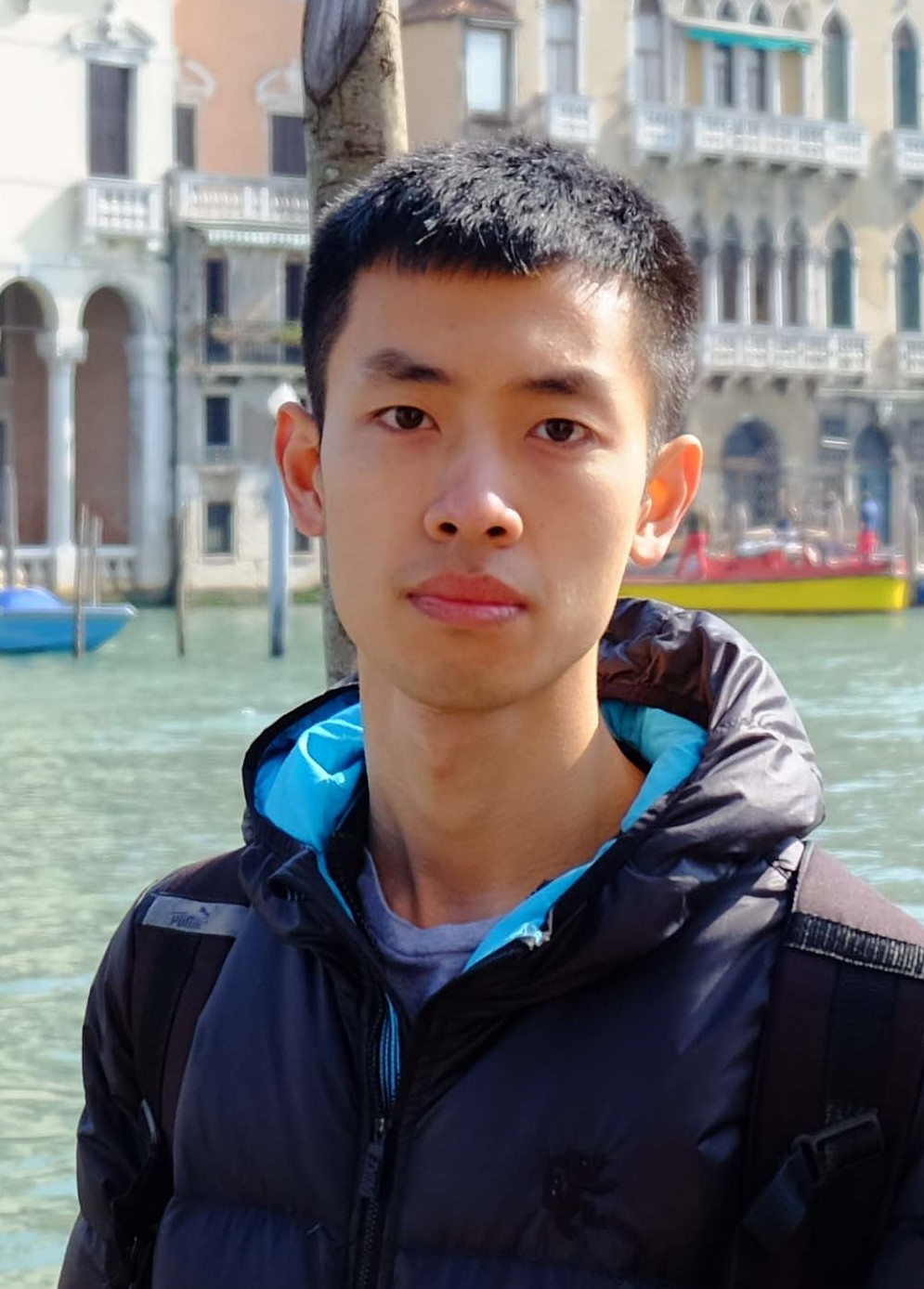}}]{Ren Li} received his B.S. (2016) in electrical engineering
  from University of Science and Technology of China in Anhui, China.
  He is currently pursuing his Ph.D. in computer engineering at Purdue
  University, IN, USA.
  His research interests include computer vision and machine learning,
  especially benefiting computer-vision systems from brain-derived
  information.
\end{IEEEbiography}

\begin{IEEEbiography}
  [{\includegraphics[width=1in,height=1.25in,clip,keepaspectratio]
      {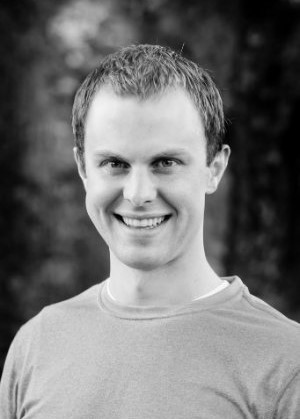}}]{Jared S. Johansen} received a B.S. in electrical
  engineering from Brigham Young University (2010) and an M.S. in electrical
  engineering and an MBA from the University of Utah (2012).
  He is currently a Ph.D. student in the School of Electrical and Computer
  Engineering at Purdue University.
  His research lies at the intersection of machine learning, computer vision,
  natural language processing, and robotics.
\end{IEEEbiography}

\begin{IEEEbiography}
  [{\includegraphics[width=1in,height=1.25in,clip,keepaspectratio]
      {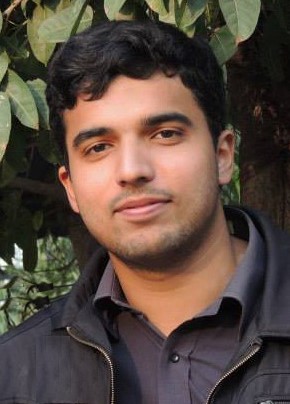}}]{Hamad Ahmed} received his B.S. in electrical engineering
  from University of Engineering and Technology Lahore, Pakistan.
  He is currently a PhD. student in the School of Electrical and Computer
  Engineering at Purdue University, IN, USA.
  His research interests lie at the intersection of machine learning and
  computer vision.
\end{IEEEbiography}

\begin{IEEEbiography}
  [{\includegraphics[width=1in,height=1.25in,clip,keepaspectratio]
      {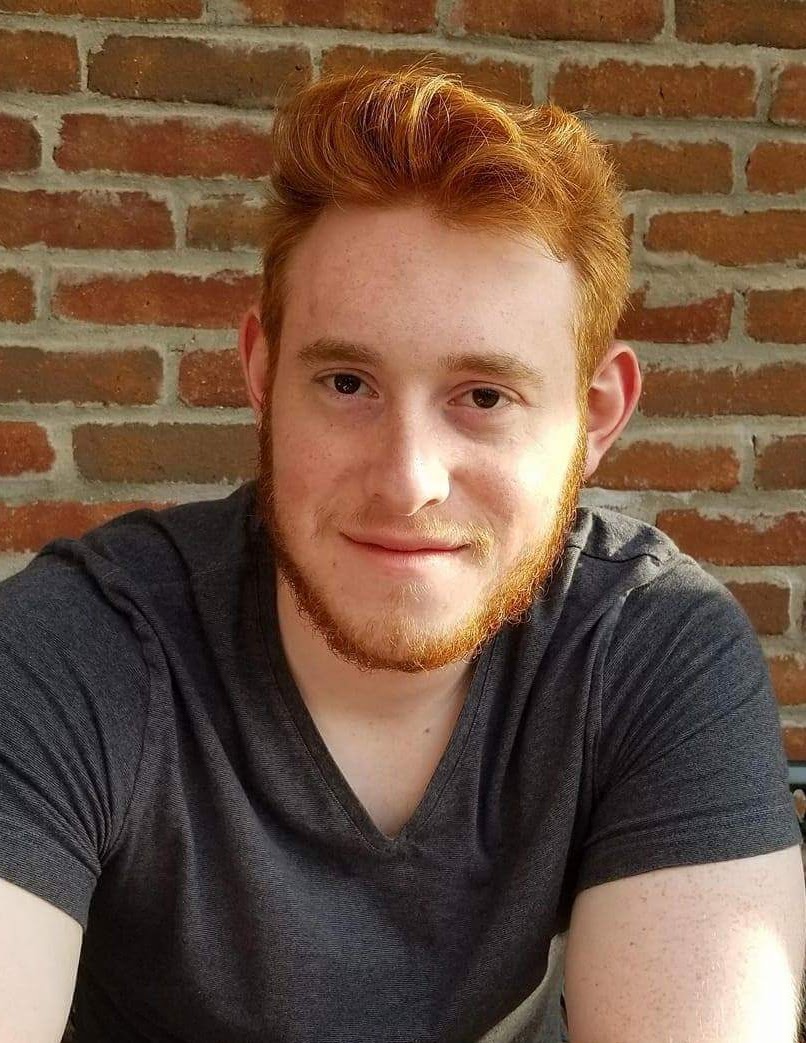}}]{Thomas V. Ilyevsky} received his B.S. in electrical
  and computer engineering from Cornell University in 2016.
  He is currently pursuing a Ph.D. in electrical and computer engineering at
  Purdue University.
  His research focuses on artificial intelligence, computer vision, and natural
  language processing in the context of human-computer interaction.
\end{IEEEbiography}

\begin{IEEEbiography}
  [{\includegraphics[width=1in,height=1.25in,clip,keepaspectratio]
      {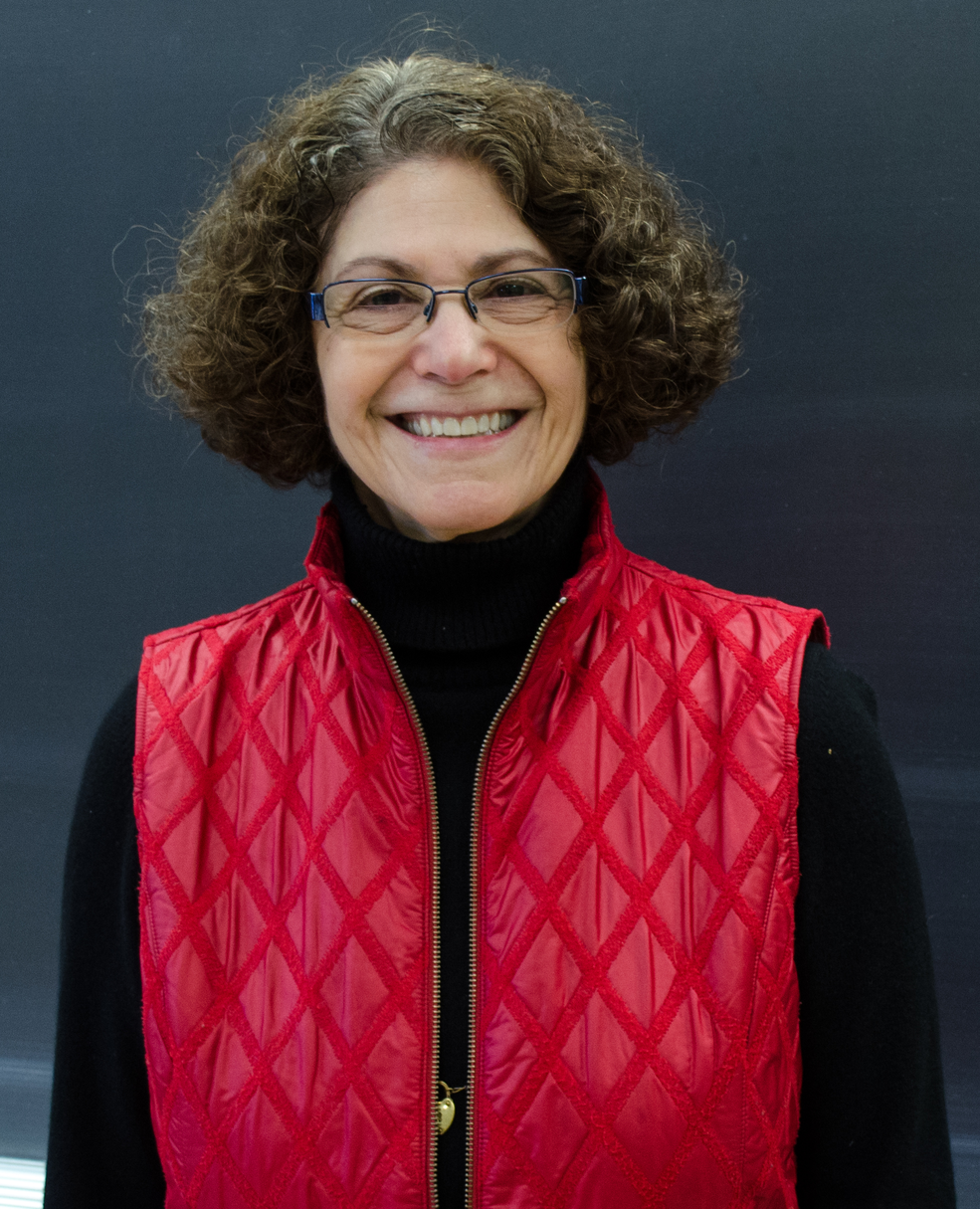}}]{Ronnie B Wilbur} has a BA (Rochester) and
  Ph.D. (UIUC) in Linguistics.
  She is Professor of Speech, Language, and Hearing Sciences in the College
  of Health and Human Sciences, and Professor of Linguistics in the College of
  Liberal Arts, at Purdue University.
  Prior to starting at Purdue in 1980, she was visiting faculty at University
  of Southern California and faculty at Boston University.
  She has been widely invited as Visiting Professor (Amsterdam, Graz, Zagreb,
  Paris ENS, Salzburg, Stuttgart) and has been funded by both the National
  Science Foundation and the National Institutes of Health.
\end{IEEEbiography}

\begin{IEEEbiography}
  [{\includegraphics[width=1in,height=1.25in,clip,keepaspectratio]
      {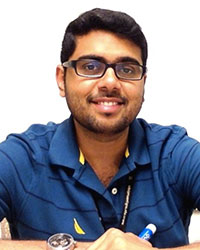}}]{Hari M Bharadwaj} is an Assistant Professor at Purdue
  University in the Department of Speech, Language, and Hearing Sciences, and
  the Weldon School of Biomedical Engineering.
  Hari received a B.Tech. in Electrical Engineering from Indian Institute of
  Technology, Madras, in 2006, and M.S. degrees in Electrical Engineering and
  Biomedical Engineering from the University of Michigan, Ann Arbor, in 2008.
  In 2014, he completed a Ph.D. in Biomedical Engineering at Boston
  University.
  Following post-doctoral work at the Martinos Center for Biomedical Imaging
  at Massachusetts General Hospital, Hari joined the faculty at Purdue
  University in 2016, where his lab integrates a multidisciplinary array of
  tools to investigate the neural mechanisms underlying auditory perception in
  humans.
\end{IEEEbiography}

\begin{IEEEbiography}
  [{\includegraphics[width=1in,height=1.25in,clip,keepaspectratio]
      {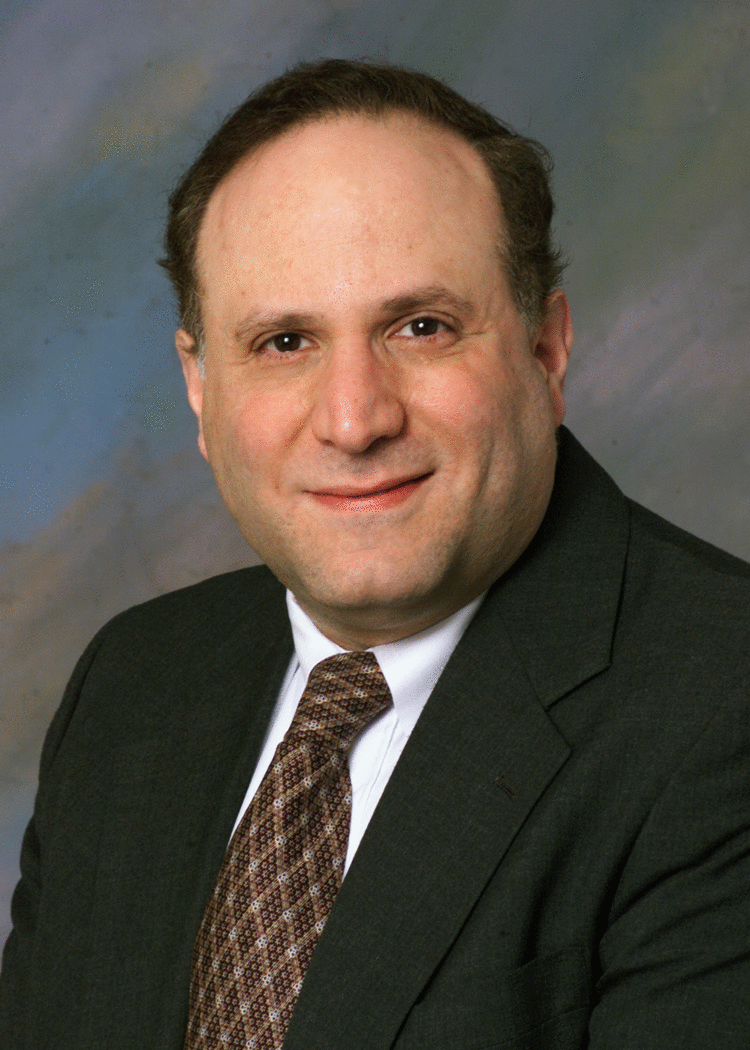}}]{Jeffrey Mark Siskind} received the B.A. degree in
  computer science from the Technion, Israel Institute of Technology in 1979,
  the S.M. degree in computer science from MIT in 1989, and the Ph.D. degree
  in computer science from MIT in 1992.
  He did a postdoctoral fellowship at the University of Pennsylvania
  Institute for Research in Cognitive Science from 1992 to 1993.
  He was an assistant professor at the University of Toronto Department of
  Computer Science from 1993 to 1995, a senior lecturer at the Technion
  Department of Electrical Engineering in 1996, a visiting assistant professor
  at the University of Vermont Department of Computer Science and Electrical
  Engineering from 1996 to 1997, and a research scientist at NEC Research
  Institute, Inc.\ from 1997 to 2001.
  He joined the Purdue University School of Electrical and Computer
  Engineering in 2002 where he is currently an associate professor.
\end{IEEEbiography}

\end{document}